%% file: main_arxiv.tex
\documentclass{article} % For LaTeX2e
\usepackage{iclr2026_conference,times}
\usepackage{graphicx}
\usepackage{hyperref}
\usepackage{amsmath,amsfonts}
\usepackage{algpseudocode}
\usepackage{array}
\usepackage{subcaption}
\usepackage{textcomp}
\usepackage{float}
\usepackage{stfloats}
\usepackage{url}
\usepackage{verbatim}
\usepackage{amssymb}
\usepackage{multirow}
\usepackage[ruled,linesnumbered]{algorithm2e}
\usepackage{booktabs}
\usepackage{colortbl}
\usepackage{color}
\usepackage{bbding}
\usepackage{wrapfig}
\usepackage{balance} 
\usepackage[table]{xcolor}
\usepackage[most]{tcolorbox} % 核心宏包
\definecolor{oursburgundy}{HTML}{DCEEFF}
\definecolor{warmback}{RGB}{254, 252, 240}   % 极浅的米色
\definecolor{warmframe}{RGB}{225, 215, 190}  % 沙色边框
\definecolor{warmtitle}{RGB}{140, 60, 40}    % 深赤褐色标题

\newtcolorbox{querybox}{
    colback=warmback,
    colframe=warmframe,
    coltext=black,
    boxrule=0.6pt,
    arc=0pt, 
    left=10pt,              % 左缩进
    right=5pt,             % 右缩进
    top=8pt,                % 上边距
    bottom=8pt,             % 下边距
    enhanced,               % 启用增强模式
    breakable               % 允许跨页（如果内容很多的话）
}

\title{Decoding Affective Nuances: Enhancing MLLMs via Hierarchical Emotion Reasoning and Contrastive Discriminative Pruning}

\author{Cheng Ye$^1$, Weidong Chen$^1$, Zhaobo Qi$^2$, Beier Zhu$^1$, Zhendong Mao$^1$ \\
$^1$University of Science and Technology of China, Hefei \\
$^2$Harbin Institute of Technology, Weihai \\
\texttt{chenweidong@ustc.edu.cn}
}

\iclrfinalcopy % Uncomment for camera-ready version, but NOT for submission.

\begin{document}

\maketitle

\begin{abstract}
While multimodal large language models (MLLMs) have demonstrated exceptional capabilities in objective understanding tasks, their performance in affective reasoning still falls significantly short of human standards. We attribute it to a central capability gap: MLLMs are difficult to reliably distinguish semantically proximal emotions based on fine-grained visual evidence, which could be decoupled as two limitations: 1) \textbf{Insufficient Attribution.} The global reasoning paradigm of conventional MLLMs severely dilutes fine-grained emotion cues, where subtle emotional states are usually implicitly encoded, thereby generating emotional misjudgments in complex scenarios. 2) \textbf{Insufficient Discrimination.} Existing methods could only identify regions generally associated with emotions, which fails to distinguish discriminative regions between semantically similar emotions, leading to ambiguous emotion judgements. To overcome these limitations, we present a training-free inference-time optimization framework, named Decoding Affective Nuances (DAN). Specifically, we propose a Hierarchical Emotional Reasoning Chain (HERC) that enhances the insufficient attribution by harmonizing fine-grained scene/object-level cues and performing a soft-gated reasoning. Furthermore, to discriminate between semantically proximal emotions, we design a Contrastive Discriminative Visual Pruning (CDVP), which isolates discriminative visual tokens to reason the final emotion category by computing the absolute discrepancy between the attention distributions of similar emotions. Performances on several benchmarks demonstrate that DAN significantly improves discrimination for affective nuances without consuming additional training resources, especially achieving +10.47\% improvements with Qwen3-VL-8B-Instruct on WebEmo25 dataset that contains 25 fine-grained emotion categories.\footnote{Code will be released in the final version of the paper.}
\end{abstract}

\input{Sec/intro}
\input{Sec/method}

\input{Sec/exp}

\input{Sec/conclu}

\section*{AI Use Statement}
In this work, we used generative AI tools for polishing the language of the manuscript and for writing and debugging portions of the experiment code. We have not used generative AI tools for the motivation or the production of the reported results and figures. These are carried out directly by the authors. All AI-assisted code was tested and verified by running the experiments reported in this paper, and the resulting claims were checked against the run outputs by the authors. We take responsibility for the final content of this work, including the text, claims, and artifacts produced with the aid of generative AI.

\bibliography{iclr2026_conference}
\bibliographystyle{iclr2026_conference}

\clearpage

 \appendix
 \section{Appendix}

\input{Sec/appendix}

\end{document}

%% file: Sec/intro.tex
\section{Introduction}

% Multimodal Large Language Models (MLLMs) have garnered widespread attention within the community due to their powerful reasoning and generation capabilities ~\citep{fei2024video,ye2024mplug,mitra2024compositional}. Driven by a wealth of related research, they have demonstrated exceptional performance across a multitude of general tasks, ranging from visual grounding ~\citep{xu2025mc,zhang2024groundhog} to complex visual understanding and reasoning ~\citep{zheng2023ddcot,yao2025r1}. By aligning high-resolution visual features with the vast knowledge base of Large Language Models (LLMs), MLLMs could achieve powerful cross-modal alignment and fusion, and be applied to a diverse range of downstream tasks. However, when transitioned from general understanding to emotion understanding, MLLMs still encounter significant bottlenecks. Emotion understanding demands more than just identifying objective facts. It requires a nuanced interpretation of complex emotion cues, where subtle environmental atmospheres and localized expressions must be synergized to form a comprehensive and accurate emotional judgment.

Multimodal Large Language Models (MLLMs) have garnered widespread attention within the community due to their powerful reasoning and generation capabilities ~\citep{fei2024video,ye2024mplug,mitra2024compositional,ye2024dual}. They have demonstrated exceptional performance across general tasks. However, when transitioning from general to emotion understanding, MLLMs still encounter significant bottlenecks.  Unlike objective recognition, which is often supported by explicit visual entities and attributes, emotion understanding requires the model to integrate contextual information with subtle affective cues, such as facial expressions, body postures, and interactions among subjects. This requirement becomes particularly challenging when the candidate labels are semantically proximal and can only be distinguished through small but decisive visual differences.

% Emotion understanding demands more than just identifying objective facts. It requires a nuanced interpretation of complex emotion cues, where subtle environmental atmospheres and localized expressions must be synergized to form a comprehensive and accurate emotional judgment.

\begin{figure*}[t]        
\center{\includegraphics[width=1.0\linewidth] {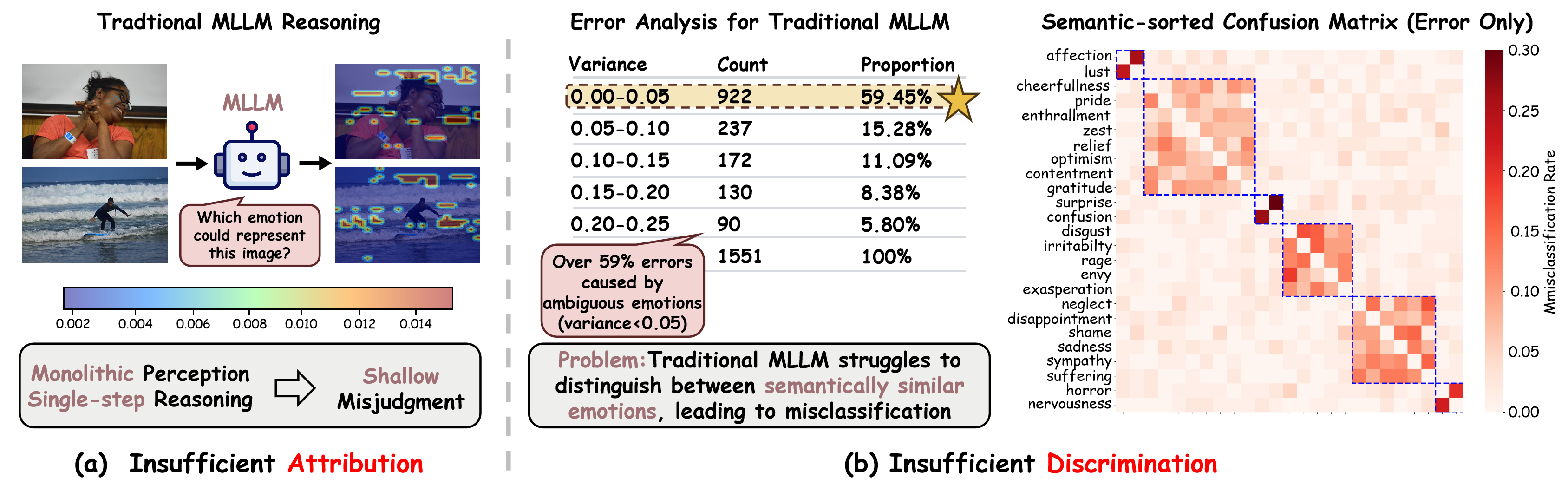}} 
\caption{Two coupled limitations underlying the difficulty of MLLMs in distinguishing semantically proximal emotions using fine-grained visual evidence.  (a) Insufficient Attribution: Global and undifferentiated visual processing dilutes localized affective cues. (b) Insufficient Discrimination: Failure cases exhibit small confidence margins and concentrated confusion among emotions within the same semantic macro-category, indicating difficulty in resolving ambiguous candidates.}
\label{fig1}
\vspace{-15pt}
\end{figure*}

% In recent years, some studies have attempted to enhance the capabilities of MLLMs in emotion understanding ~\citep{cheng2024emotion,fang2025emoe}. Specifically, some of them focus on constructing high-quality emotional datasets including fine-grained annotations and diverse instances, aiming to enhance emotion understanding capabilities of MLLMs through fine-tuning ~\citep{lian2025affectgpt,lian2025ov}. Additionally, other studies introduce specialized adapters to facilitate the fusion of multimodal emotional information ~\citep{yang2025mse}. However, these works bring substantial overhead in both manual annotations and training costs, limiting their scalability and efficiency. Therefore, some studies have shifted toward training-free paradigms ~\citep{zhang2024visual}. They rely on auxiliary annotations for images to guide MLLMs focus on emotion-related regions, which still necessitates additional human annotations. 

% In this paper, we attempt to enhance the emotional comprehension of MLLMs during the inference-time through sophisticated prompt engineering and discriminative visual token pruning. Without any additional training costs or external annotations, our framework significantly augments the scalability and efficiency of models in real-world scenarios.
In recent years, some studies have attempted to enhance the capabilities of MLLMs in emotion understanding ~\citep{cheng2024emotion,fang2025emoe,huang2025graph,chen2021cascade}. Specifically, some of them focus on constructing high-quality emotional datasets and introducing specialized fusion adapters to fine-tune the MLLMs ~\citep{lian2025affectgpt,lian2025ov,yang2025mse,chen2022multi}. However, these works bring substantial overhead in both manual annotations and training costs, limiting their scalability and efficiency. Therefore, some studies have shifted toward training-free paradigms ~\citep{zhang2024visual,chen2023weakly,chen2026creatiparser}. They attempt to guide MLLMs to perform emotion reasoning by exploring the attention distribution on visual information through prompt engineering during the inference phase. 
Despite credible progress, existing methods still struggle with a central capability gap: \textbf{MLLMs are difficult to reliably distinguish semantically proximal emotions based on fine-grained visual evidence.} We attribute this capability gap to two crucial limitations. 
1) \textbf{Insufficient Attribution.} Traditional MLLMs rely on global visual processing and directly predict an emotion label. Such global reasoning severely dilutes fine-grained cues, \emph{i.e.,} facial expressions and body postures, where subtle emotional states are usually implicitly encoded. As shown in Fig. 1(a), much irrelevant background information is erroneously highlighted, while key emotion-related regions are overshadowed, which ultimately leads to emotion misjudgments.
2) \textbf{Insufficient Discrimination.} Secondly, obtaining fine-grained evidence is not equivalent to achieving fine-grained discrimination. Semantically proximal emotions often share partial visual evidence. Existing pruning-based methods~\citep{fang2025catch} could only identify regions generally associated with emotions, which fails to reveal which regions tend to support one candidate over another. Such a mechanism inherently overlooks the fine-grained boundary delineation between semantically similar emotions. 
As shown in Fig. 1(b), we conduct statistical analysis for failure cases of the baseline model~\citep{fang2025catch} on the WebEmo25 dataset~\citep{panda2018contemplating}\footnote{More details of variance analysis are shown in the Appendix.}. We calculate the confidence variance between the top-2 predicted emotions for failure cases and observe that over 59\% of the failure cases are concentrated in the low-confidence intervals, indicating severe ambiguity during inference. Furthermore, we plot a confusion matrix correlating the ground-truth labels with the misclassified predictions. Based on Parrott's emotion theory~\citep{parrott2001emotions}, we group the 25 fine-grained emotions into 6 macro-categories and order them accordingly along the axes. We observe that the misclassifications are densely concentrated within the six highlighted diagonal sub-matrices, which correspond to semantically proximal emotions. Such insufficient discrimination for semantically similar emotions leads to ambiguous confidence margins between candidate emotions, ultimately culminating in emotional misjudgments.

To address these limitations, we propose \textbf{Decoding Affective Nuances (DAN)}, a training-free inference-time framework to sharpen the ambiguous emotion discrimination through two successive modules. Specifically, we first introduce a {Hierarchical Emotion Reasoning Chain (HERC)} to improve fine-grained affective evidence acquisition by explicitly mining complementary scene-level and object-level cues. It subsequently performs soft-gated reasoning from emotion polarity to fine-grained categories, producing a stable preliminary distribution while reducing error propagation from hard intermediate decisions. 
Subsequently, we propose the {Contrastive Discriminative Visual Pruning (CDVP)} module to perform evidence discrimination for ambiguous samples. CDVP first constructs an ambiguous emotion set and obtains candidate-specific attention maps through contrastive prompts. Regions with large attention discrepancies provide stronger evidence for distinguishing candidates. CDVP therefore preserves high-discrepancy visual tokens and prompts the MLLM to reselect the final emotion based on the discriminative evidence.
Experimental evaluations on several public benchmarks demonstrate that our proposed DAN framework significantly outperforms the state-of-the-art methods in emotion recognition. Notably, our framework achieves these gains without any additional training or annotations, offering a highly scalable and resource-efficient solution. In summary, our contributions are:

$\bullet$ We identify a capability gap in MLLM-based emotion understanding: the difficulty of distinguishing semantically proximal emotions using fine-grained visual evidence. We further characterize this gap through two limitations, namely insufficient attribution and insufficient discrimination.

$\bullet$ We propose Decoding Affective Nuances (DAN), a training-free inference-time framework that addresses two limitations through complementary modules. HERC acquires scene-level and object-level affective evidence to produce a stable preliminary emotion distribution, while CDVP explicitly compares ambiguous candidates and preserves discriminative visual tokens for final reselection.

$\bullet$ Extensive experiments on several public benchmarks prove the effectiveness of our framework and each proposed module. Especially on WebEmo25 dataset that contains 25 fine-grained emotion categories, DAN achieves +10.47\% improvements with Qwen3-VL-8B-Instruct on emotion accuracy.

%% file: Sec/method.tex
\section{methodology}
\subsection{Preliminary}
\begin{figure*}[t]        
\center{\includegraphics[width=0.95\linewidth] {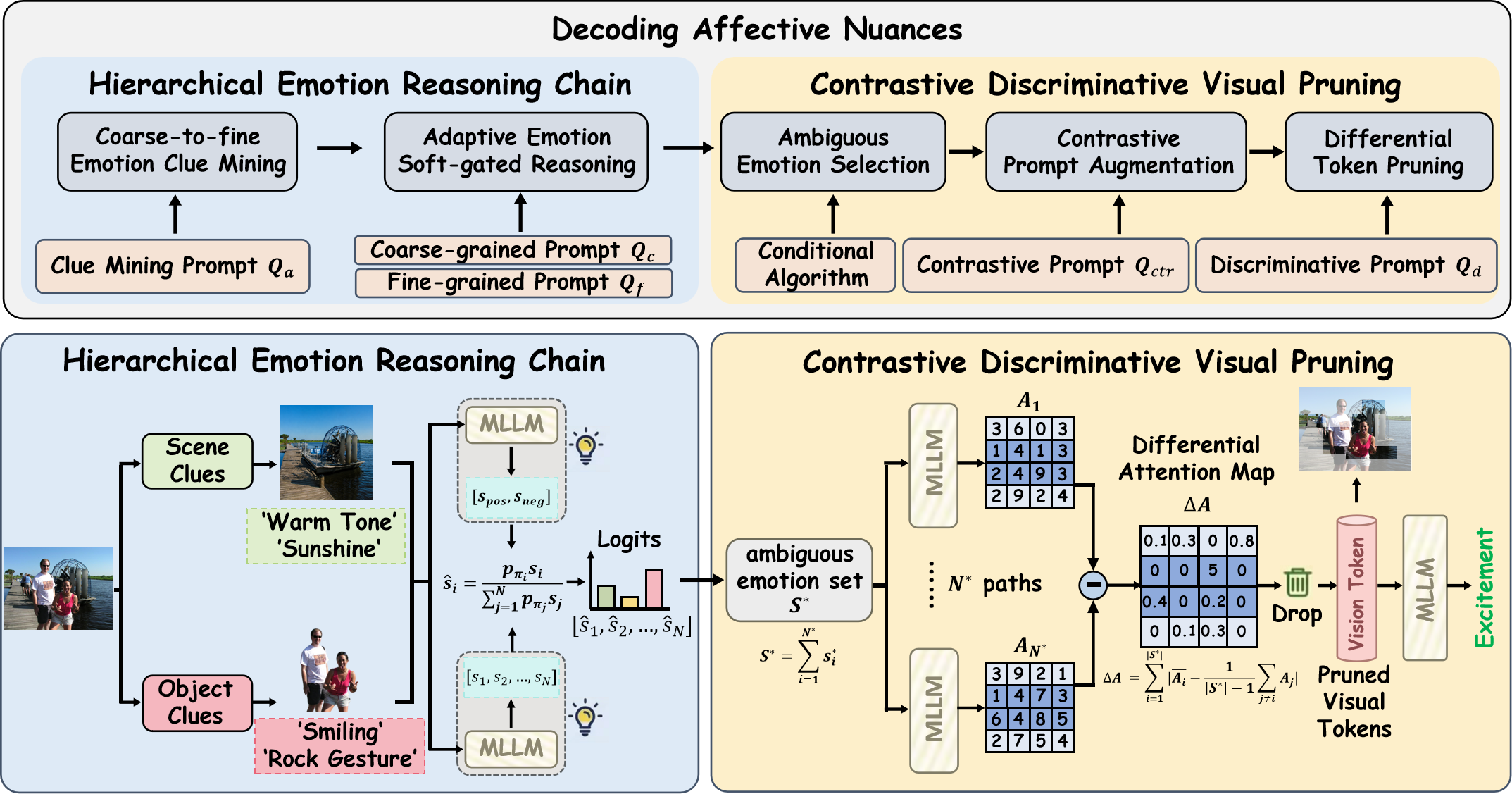}} 
\caption{The illustration of our proposed DAN framework. The Hierarchical Emotion Reasoning Chain module (HERC) firstly identifies the scene-level and object-level emotion clues, and then performs a soft-gated reasoning for a stable preliminary classification. Then, the Contrastive Discriminative Visual Pruning module (CDVP) is proposed to handle the affectively nuanced samples, which guides MLLM to focus on the most discriminative regions and reselect the final emotion.}
\label{fig2}
\vspace{-15pt}
\end{figure*}

In this paper, we propose Decoding Affective Nuances (DAN), an inference-time training-free framework designed to enhance the emotion understanding of MLLMs, especially for confusing and ambiguous emotional scenarios. The model processes an input pair consisting of a visual image $\mathcal{V}$ and a textual prompt $\mathcal{Q}$ to generate a descriptive response that categorizes the emotion. As shown in Fig. \ref{fig2}, our proposed DAN mainly consists of two primary components: i) Hierarchical Emotion Reasoning Chain (HERC) that captures the preliminary emotion ranking and confidence scores, and ii) Contrastive Discriminative Visual Pruning (CDVP) that leverages the preliminary distribution to select ambiguous samples and emotions and then performs discriminative reselection.

\subsection{Hierarchical Emotion Reasoning Chain}\label{HERC}
\textbf{Motivation.} Given the complex and cue-dependent nature of emotions, emotional states are usually embedded in fine-grained visual cues that are difficult for global visual reasoning to detect~\citep{weng2023affective,you2016building,ye2025multi,ye2025improving}. Therefore, a fine-grained reasoning process ensures precise and logically grounded emotion reasoning. Inspired by it, we propose the Hierarchical Emotion Reasoning Chain (HERC) to empower MLLMs to focus on fine-grained affective cues during the inference phase.

% \begin{wrapfigure}{r}{0.55\textwidth} % r 表示居右，占 0.5 宽度
%   \centering
%   \vspace{-15pt}
%     \begin{querybox}
%     {\textbf{\textcolor{warmtitle}{Clue Mining Prompt $\mathcal{Q}_a$:}}} \\
%     \textit{Please identify potential affective-arousing clues in the image following these steps:}\\
%     \textit{\textbf{Step 1}: Identify scene-level clues that reflect emotions (e.g., lighting and colors).}\\
%     \textit{\textbf{Step 2}: Identify subject-level clues that reflect emotions (e.g., facial and body expressions).}
% \end{querybox}
% \vspace{-15pt}
% \end{wrapfigure}

\textbf{Coarse-to-fine Emotion Clue Mining.} Firstly, we design a prompt to mine affective clues from global scenes to local objects. When identifying an image, people usually first observe the global scene to capture the overall emotional tone (\emph{e.g.}, lighting, color temperature). Subsequently, they further focus on fine-grained objects to encode crucial emotion evidence(\emph{e.g.}, facial expressions, bodily gestures). Thus, we guide MLLMs to identify two types of affective-arousing clues, namely scene-level and object-level clues through the clue mining prompt $\mathcal{Q}_a$\footnote{Details of $\mathcal{Q}_a$ and the subsequently mentioned prompts are all presented in the Appendix.}.

% \begin{querybox}
%     {\textbf{\textcolor{warmtitle}{Attribute Mining Prompt $\mathcal{Q}_a$:}}} \\
%     \textit{Please identify potential affective-arousing attributes in the image following these steps:}\\
%     \textit{\textbf{Step 1}: Identify scene-level attributes that reflect emotions (e.g., lighting and colors).}\\
%     \textit{\textbf{Step 2}: Identify subject-level attributes that reflect emotions (e.g., facial and body expressions).}
% \end{querybox}

% In the second stage, we constrain MLLMs to focus on the concrete expressions of these attributes. For instance, warm sunlight or an upturned mouth serves as explicit visual evidence pointing toward positive affective states. The specific prompt $\mathcal{Q}_e$ is shown above.

\textbf{Adaptive Emotion Soft-gated Reasoning.} Based on the above affective clue mining, we introduce an adaptive soft-gated reasoning strategy to achieve a multi-granular emotion reasoning. Specifically, the entire emotion word set $S=\{e_i\}_{i=1}^{N}$ is divided into the positive emotion subset $[{\rm PES}]$ and the negative emotion subset $[{\rm NES}]$, where $N$ is the number of emotion category. We first design a coarse-grained prompt $\mathcal{Q}_c$ to determine the emotional polarity of the image. By combining $\mathcal{Q}_a$ and $\mathcal{Q}_c$, the MLLM could generate the emotion polarity label. However, to mitigate the impact of error propagation during hierarchical reasoning, instead of directly employing hard labels, we generate soft polarity scores by the logits generator of MLLM as follows:
\begin{equation}
    p_{\rm pos},p_{\rm neg} = {\rm Softmax}({\rm LogGen}_{\mathcal{M}}([\mathcal{Q}_a, \mathcal{Q}_c],\mathcal{V})),
\end{equation}
{where ${\rm LogGen},\mathcal{M},\mathcal{V}$ denote the logits generator, MLLM, and visual image, respectively. Subsequently, in the fine-grained stage, we modify a fine-grained prompt $\mathcal{Q}_f$ to classify specific emotion categories. Similarly, combined $\mathcal{Q}_a$ and $\mathcal{Q}_f$, we generate soft fine-grained scores for each emotion word by the logits generator of MLLM:}
\begin{equation}
    [s_{1},s_2,\cdots,s_{N}] = {\rm Softmax}({\rm LogGen}_{\mathcal{M}}([\mathcal{Q}_a, \mathcal{Q}_f],\mathcal{V})),
\end{equation}
Finally, the aggregated scores for each emotion word are computed by the combination of the polarity score and fine-grained score:
\begin{equation}
    \hat{s}_{i} = \frac{p_{\pi_{i}}s_i}{\sum_{j=1}^N p_{\pi_j}s_j},
\end{equation}
where $\pi_i\in\{pos,neg\},\hat{S}=[\hat{s}_{1},\hat{s}_2,\cdots,\hat{s}_{N}]$ denote the polarity of $i$-th emotion word and entire probability distribution. The proposed HERC constraints MLLMs to generate preliminary emotion predictions by coarse-to-fine clue mining and adaptive soft-gated calculation, mitigating the error accumulation in hierarchical reasoning and providing a stable anchor for discriminative pruning.
% \begin{querybox}
%     {\textbf{\textcolor{warmtitle}{Coarse-grained Prompt $\mathcal{Q}_c$:}}} \\
%     \textit{Based on the above analysis, choose an option that best represents the image:}
%     \begin{center}
%         \textbf{1}. Positive \qquad \textbf{2}. Negative
%     \end{center}
%     \textit{Positive emotions include $[{\rm PES}]$. Negative emotions include $[{\rm NES}]$. Answer directly with the number of the chosen option.}
% \end{querybox}

 % \begin{querybox}
 %     {\textbf{\textcolor{warmtitle}{Fine-grained Prompt $\mathcal{Q}_f$:}}} \\
 %     \textit{Based on the above analysis, choose an option that best represents the image:}
 %     \begin{center}
 %         \textbf{1}. $e_1$ \qquad \textbf{2}. $e_2$ \qquad $\cdots$ \qquad \textbf{N}. $e_N$
 %     \end{center}
 %     \textit{Answer directly with the number of the chosen option.}
 % \end{querybox}

\subsection{Contrastive Discriminative Visual Pruning}\label{CDVP}
\textbf{Motivation.} Based on statistical analysis, we observe a significant positive correlation signal between the confidence variance of semantically proximal emotional categories and the overall classification accuracy. Meanwhile, rather than executing an open selection among confusing options, we analyze that it is inherently more manageable to perform a contrastive assessment. Consequently, we design the Contrastive Discriminative Visual Pruning (CDVP) module. By constructing contrastive prompts and performing differential attention computations, CDVP constraints the MLLM to focus on the most discriminative visual regions, thereby significantly enhancing reasoning precision.  

\textbf{Conditional Ambiguous Emotion Selection.} Given the limited capability of MLLMs to discriminate between semantically similar emotions, we make a statistical analysis to find that the error rate improves significantly on hard samples that easily induce emotional confusion. To address this, we design a conditional ambiguous emotion selection algorithm to determine whether a sample is a hard sample by selecting the corresponding ambiguous emotion set $S^*$. 
Specifically, we sort $\hat{S}$ in descending order and calculate the variance between the top-1 score $s_{1}$ and each score. If the variance is less than a predefined threshold $\alpha$, these emotions are deemed ambiguous and are divided into $S^*$. This iterative process continues until a variance exceeding the threshold. Besides, we set a truncation threshold $k$ to balance the computational overhead. Meanwhile, we argue that the confidence of lower-ranked emotion scores diminishes progressively and contains noise. The overall procedure is summarized in Algorithm 1. 

\begin{wrapfigure}{r}{0.5\textwidth} % {r} 表示靠右，0.5\textwidth 表示占据一半宽度
  \begin{minipage}{0.5\textwidth}
    \begin{algorithm}[H] % 注意这里必须用 [H]，强制固定位置
      \caption{Conditional Ambiguous Emotion Selection}
      \label{alg:hard_sample}
\KwIn{Descending Emotion Probability Distribution $\hat{S}$,
Variance Threshold $\alpha$, 
Truncation Threshold $k$, Top-1 Score $s_{1}$}
\KwOut{Ambiguous emotion set $S^*$.}
\BlankLine
\textbf{Initialize:}$S^* \leftarrow s_{1}$ \\
\For{$i = 2, \dots, \min(k+1, |S|-1)$}{
    $\mu_i \leftarrow (s_i + s_{max}) / 2$ \;
    
    $\sigma^2_i \leftarrow \left[(s_i - \mu_i)^2 + (s_{max} - \mu_i)^2\right] / 2$
    
    \If{$\sigma^2_i < \alpha$}{
        $S^* \leftarrow S^* \cup \{s_i\}$ 
    }
    \Else{
        \textbf{break} 
    }
}
\Return{Ambiguous emotion set $S^*$.}
\end{algorithm}
  \end{minipage}
\end{wrapfigure}

Finally, for samples with $|S^*|=1$, we argue that the prediction of HERC is highly confident, and directly output the emotion category with the highest score as the final prediction. Conversely, this sample will be classified as a hard sample and fed into the CDVP module for further discriminative re-selection:
\begin{equation}
    \mathcal{E} = \begin{cases} e_{{\rm argmax}(\hat{S})}, & |S^*|=1, \\ {\rm CDVP}(S^*), & |S^*|\geq 2. \end{cases}
\end{equation}
This adaptive mechanism ensures that computational resources are strategically allocated to resolve the most challenging and affectively nuanced cases.

% \begin{querybox}
%     {\textbf{\textcolor{warmtitle}{Contrastive Prompt $\mathcal{Q}_{ctr}(s^*_i,S^*)$:}}} \\
%     \textit{Please analyze that why should this image be categorized as {$s^*_i$} rather than {$S^*\setminus s^*_i$}? Please pinpoint the visual cues that support this distinction.}
% \end{querybox}
\textbf{Contrastive Discriminative Pruning.} Guided by the conditional ambiguous emotion selection, we isolate challenging samples with significant emotional ambiguity. Subsequently, we introduce a contrastive prompting strategy to bolster affective reasoning. Rather than forcing the MLLM to make a selection among the emotion set, we employ a contrastive enhancement mechanism that compels MLLM to explain the discriminative visual evidence justifying a specific emotional category over its competitors. Guided by the contrastive prompt $\mathcal{Q}_{ctr}$, MLLM no longer needs to make simple choices merely. It also needs to provide concrete visual evidence, which forces MLLM to shift from global perception to local mining, precisely focusing on discriminative visual regions closely related to the current emotion. Subsequently, to select the visual regions that best distinguish the ambiguous emotions, we employ contrastive prompts to guide MLLM to focus on each ambiguous emotion, respectively. Following this, a natural strategy is to extract the attention map from the average of the last-3 transformer layers of MLLM to quantify the significance of each visual token relative to the specific emotion. Specifically, we first extract the attention map from MLLM, and compute the average score of all textual tokens to quantify the significance of each visual token:
\begin{align}
    \mathcal{A}_i = {\rm Attn}_{\mathcal{M}}(\mathcal{Q}_{ctr}(s^*_i,S^*),\mathcal{V})\in\mathbb{R}^{N_t\times N_v},\\
    \overline{\mathcal{A}}_i = {\rm Avg}_{\rm T-axis}(\mathcal{A}_i)\in\mathbb{R}^{N_v}, i=1\ to\ |S^*|
\end{align}
where $N_t,N_v$ denote the number of textual and visual tokens, respectively. Subsequently, we eliminate common redundancy and select the discriminative visual regions by a joint discrepancy computation. Specifically, we quantify the significance of visual tokens by computing the mean absolute discrepancy between each ambiguous emotion and its competitors. Under this mechanism, only tokens that significantly distinguish a particular category from all other categories will receive high scores:
\begin{equation}
    \Delta \mathcal{A} = \sum_{i=1}^{|S^*|} |\overline{\mathcal{A}}_i - \frac{1}{|S^*|-1} \sum_{j \neq i} \overline{\mathcal{A}_j}|
\end{equation}
In the differential attention map $\Delta\mathcal{A}$, we select low-ranking tokens to generate a dropping mask $\mathcal{D}$, which guides MLLM to focus on the discriminative visual regions:
\begin{equation}
    \mathcal{D} = \{{a}_i|i\in {\rm argmin_G(\Delta \mathcal{A})}\},\quad G=\lfloor \beta N_v \rfloor
\end{equation}
where $\beta$ is the dropping ratio. Finally, based on the dropping mask, we design a discriminative prompt $\mathcal{Q}_d$ to select the final emotion category from $S^*$. We apply $\mathcal{Q}_d$ along with the pruned image into the MLLM for final selection:
\begin{align}
    &\mathcal{V}'=\{v_i|v_i\in (\mathcal{V-\mathcal{D})}\},\\
    &\mathcal{E} = \mathcal{M}(\mathcal{Q}_d,\mathcal{V}'),
\end{align}
guided by the contrastive discriminative visual pruning, MLLM focuses on the visual regions that are most capable of distinguishing ambiguous emotions and correct potential initial misjudgments, which is crucial for more accurate and reliable emotion classification.

% \begin{querybox}
%     {\textbf{\textcolor{warmtitle}{Discriminative Prompt $\mathcal{Q}_{d}$}}} \\
%     \textit{Based on the discriminative visual image, choose an option that best represents the image:}
%     \begin{center}
%         \textbf{1}. $s^*_1$ \qquad \textbf{2}. $s^*_2$ \qquad $\cdots$ \qquad \textbf{N$^*$}. $s^*_{N^*}$
%     \end{center}
%     \textit{Answer directly with the number of the chosen option.}
% \end{querybox}

%% file: Sec/exp.tex
\begin{table*}[t]
\centering
\setlength{\aboverulesep}{0pt}
\setlength{\belowrulesep}{0pt}
\caption{\label{main}{{Comparison with state-of-the-art on various emotion datasets. The optimal results are denoted by boldface.}}} 
\scalebox{0.9}{
\begin{tabular}{l|ccccc|c}
\toprule
 \rowcolor{gray!40}{Dataset}  &  Emotion6 & EmoSet8 & WebEmo7 & WebEmo25 & Abstract8 & Average \\ 
\midrule
\multicolumn{7}{c}{\textit{Qwen2.5-VL-7B-Instruct}}  \\ 
\cmidrule{1-7} 
\rowcolor{gray!15}Zero-shot & {58.33}  &  56.98  & {47.70} & {22.45} & {23.68} &41.83 \\
Zero-shot-CoT &  {{59.76}}   & {{57.19}}    & {47.25} &{22.15}  & {23.25} & 41.92\\
SEPM &  \underline{61.58 }& \underline{57.94}  & \underline{49.10} & \underline{22.85}  & \underline{26.32} &\underline{43.56} \\
\rowcolor{oursburgundy}\textbf{DAN(Ours)}   &\textbf{63.13} &\textbf{59.10} & \textbf{50.90} & \textbf{24.85} &\textbf{29.82}  & \textbf{45.56} \\
\cmidrule{1-7}
\multicolumn{7}{c}{\textit{Qwen3-VL-4B-Instruct}}  \\ 
\cmidrule{1-7}
\rowcolor{gray!15}Zero-shot & 55.21 & 57.06 & 48.60 & 22.40 & 20.18 &40.69 \\
Zero-shot-CoT & 56.22 & 59.26 & 47.70 &22.45  & 20.00 &41.13 \\
SEPM & \underline{ 60.31} & \underline{63.28 } & \underline{50.25 } & \underline{ 23.05} & \underline{23.68 }& \underline{44.11 } \\
\rowcolor{oursburgundy}\textbf{DAN(Ours)}& \textbf{62.29 }  & \textbf{65.66 }  & \textbf{52.05 }  & \textbf{24.95 } & \textbf{26.75 }  &  \textbf{46.34 }\\
\cmidrule{1-7}
\multicolumn{7}{c}{\textit{Qwen3-VL-8B-Instruct}}  \\ 
\cmidrule{1-7} 
\rowcolor{gray!15}Zero-shot & 55.21 & 56.50 & 48.00 & 21.10 & 29.38 &42.04 \\
Zero-shot-CoT & 52.85 & 55.18 & 48.75 & 21.75 & 28.93 &41.49 \\
SEPM & \underline{55.97 } & \underline{57.25 } & \underline{51.30 } & \underline{22.45 } & \underline{ 30.70}& \underline{43.53 } \\
\rowcolor{oursburgundy}\textbf{DAN(Ours)}& \textbf{59.09 }  & \textbf{60.21 }  & \textbf{53.40 }  & \textbf{24.80 } & \textbf{33.77 }  &  \textbf{46.25 }\\
\cmidrule{1-7}
\multicolumn{7}{c}{\textit{InternVL3.5-8B}}  \\ 
\cmidrule{1-7}
\rowcolor{gray!15}Zero-shot & 54.37 & 55.81 & 42.55 &14.45  & 26.79 &38.79 \\
Zero-shot-CoT & 54.87 & 56.49 & 42.35 & 14.95 & 30.36 &39.80 \\
SEPM & \underline{55.55 } & \underline{57.53 } & \underline{44.05 } & \underline{18.15 } & \underline{32.89 }& \underline{41.63 } \\
\rowcolor{oursburgundy}\textbf{DAN(Ours)}& \textbf{58.59 }  & \textbf{60.96 }  & \textbf{45.95 }  & \textbf{21.30 } & \textbf{35.09 }  &  \textbf{44.38 }\\
\bottomrule
\end{tabular}}
\end{table*}

\section{experiment}
\subsection{Experimental Setup}
\textbf{Dataset.} Follow the existing work settings~\citep{fang2025catch,fang2026emo,chen2026subjective,song2025towards}, we evaluate the performance of our framework on four public emotion recognition benchmarks, EmoSet~\citep{yang2023emoset}, WebEmo~\citep{panda2018contemplating}, Emotion6~\citep{peng2015mixed}, and Abstract~\citep{machajdik2010affective}. The EmoSet dataset contains $\sim$118k images and are annotated with 8 basic emotions. The WebEmo dataset contains $\sim$2k images collected from the web and includes emotion labels at two granularity levels with 7 and 25 categories. The Emotion6 dataset contains 1980 images and are annotated with 6 basic emotions. The Abstract dataset contains 228 abstract paintings labeled with 8 basic emotions through manual voting. Besides, we leverage emotion accuracy (Acc) as the evaluation metric for all experiments.

\textbf{Implementation Details.} We leverage four MLLMs to evaluate the effectiveness of our framework, including Qwen2.5-VL-7B-Instruct~\citep{qwen2025qwen25technicalreport}, Qwen3-VL-4B-Instruct~\citep{bai2025qwen3}, Qwen3-VL-8B-Instruct~\citep{bai2025qwen3}, and InternVL3.5-8B~\citep{wang2025internvl3}. We set the variance threshold $\alpha=0.125$, the dropping ratio $\beta=0.3$, and the truncation threshold $k=2$. Since our framework is inference-time and training-free, all experiments are only conducted on a single NVIDIA A800 GPU with 80GB of memory. Unless otherwise specified, all ablation studies are conducted on the Qwen3-VL-8B-Instruct model, which has the largest number of parameters.

\subsection{Main Comparison}\label{4.2}

Firstly, compared to the Zero-shot approach\footnote{Details of baseline methods are shown in the Appendix.}, our model yields a significant improvement in the average accuracy, \emph{i.e.,} +8.8\%/+13.8\%/+10.0\%/+{14.4}\% with all four MLLMs, respectively. While conventional MLLMs excel in objective comprehension tasks, they encounter a critical bottleneck in subjective emotion understanding. Our proposed DAN framework effectively enhances the emotional sensitivity of MLLMs during the inference phase.
Besides, our model also outperforms the Zero-shot-CoT method in the average accuracy, \emph{i.e.,} +8.7\%/+11.6\%/+11.5\%/+{11.5}\% with all four MLLMs, respectively. We observe that the traditional Chain-of-Thought paradigm, characterized by a single-step sequential reasoning pattern, could be negative for emotion-related tasks, leading to a performance degradation relative to the Zero-shot baseline. In contrast, our HERC module enhances hierarchical emotion inference through coarse-to-fine cue mining and soft-gated reasoning.
Finally, compared to the state-of-the-art pruning-based methods SEPM, our approach also achieves average accuracy gains, \emph{i.e.,} +3.8\%/+5.0\%/+6.2\%/+{6.6}\% for SEPM with all four MLLMs, respectively. Despite their progress in reducing visual redundancy, these methods struggle with ambiguous or semantically similar emotions. Our CDVP module addresses this gap by leveraging contrastive prompt augmentation and differential attention computation, significantly empowering the MLLM to resolve fine-grained emotional nuances with high precision. 
It is worth noting that the WebEmo25 dataset annotates 25 fine-grained and semantically proximal emotion categories. Across the five benchmark settings, our model leads all four zero-shot MLLMs on WebEmo25, \emph{i.e.,} +10.7\%/+11.4\%/+17.5\%/+{47.4}\%, respectively. This observation demonstrates that our proposed CDVP module could focus on the most discriminative visual regions and effectively decouple semantically similar emotional states.

\subsection{Ablation Study}\label{4.3}

\begin{wrapfigure}{r}{0.65\textwidth} % r 表示居右，占 0.5 宽度
  \centering
  \setlength{\aboverulesep}{0pt}
\setlength{\belowrulesep}{0pt}
  \vspace{-10pt} % 根据需要调整表格上方的间距
  \caption*{Table 2: Ablation study for key components.}
  \label{component}
  \small
\begin{tabular}{ccc|ccc}
\toprule
\rowcolor{gray!40}\multicolumn{2}{c}{HERC}  &   &    &  &   \\ 
\rowcolor{gray!40} Scene & Object &\multirow{-2}*{CDVP} &\multirow{-2}*{Emotion6} &\multirow{-2}*{WebEmo25} &\multirow{-2}*{Abstract8}  \\
\midrule
{$\times$}&{$\times$} & {$\times$} & 55.21   & 21.10  & 29.38   \\
\rowcolor{gray!15}{$\checkmark$}&{$\times$} & {$\times$} & 54.88   & 21.25  & 31.14    \\
{$\times$}&{$\checkmark$} & {$\times$} &  56.40  & 22.50  & 28.51   \\
\rowcolor{gray!15}{$\checkmark$}&{$\checkmark$} & {$\times$} & 57.24   & {23.15}  & \underline{31.58 }  \\
{$\times$}&{$\times$} & {$\checkmark$} & \underline{58.08}   & \underline{23.80}  & {30.26}  \\
\rowcolor{oursburgundy}{$\checkmark$}&{$\checkmark$} & {$\checkmark$} &  \textbf{59.09}  & \textbf{24.80}  & \textbf{33.77}  \\
\bottomrule
\end{tabular}
  \vspace{-10pt} % 调整表格下方的间距
\end{wrapfigure}

\textbf{Primary Component Discussion.} To verify the effectiveness of proposed each module, we perform an ablation study for primary components, including scene-level and object-level cue mining, and the CDVP module. As shown in Table 2, we observe that relying solely on scene-level cues may cause a negative effect on some datasets, \emph{i.e.,} Emotion6. We analyze that the reason is that an overemphasis on the global emotional atmosphere leads to the neglect of fine-grained emotional details. However, in the abstract painting dataset Abstract8, the effect is positive. Meanwhile, the effect of object-level cues is opposite between normal and abstract datasets. Given that Abstract8 consists of abstract artworks with sparse objects, the emotion is predominantly implicit in global environments. Overall, the synergistic integration of these two clues provides consistent improvements across diverse scenarios. Furthermore, the CDVP module consistently enhances accuracy across all datasets by facilitating discriminative selection among semantically proximal emotions through contrastive prompting and differential pruning. These observations demonstrate the crucial role of each proposed component in boosting performance.

\begin{wrapfigure}{r}{0.55\textwidth} 
  \centering
          \setlength{\aboverulesep}{0pt}
\setlength{\belowrulesep}{0pt}
  \vspace{-10pt}
  \caption*{Table 3: Discussion on soft-gated reasoning mechanism.}
  \label{soft-gated}
  \small
\begin{tabular}{l|ccc}
\toprule
\rowcolor{gray!40}{Setting}  &  {Emotion6}   & {WebEmo25} & {Abstract8}  \\  
\midrule
\rowcolor{gray!15}No-gating & 54.55   & 20.90  &  28.51 \\
Hard-gating &  \underline{58.42}  & \underline{24.05}   &  \underline{32.02}    \\
\rowcolor{oursburgundy}Soft-gating& \textbf{59.09}  & \textbf{24.80} & \textbf{33.77} \\
\bottomrule
\end{tabular}
  \vspace{-10pt} 
\end{wrapfigure}

\textbf{Discussion on Soft-gated Reasoning.} To verify the superiority of soft-gating mechanisms, we make a comparison with hard-gating and no-gating mechanisms. Specifically, hard-gating refers to directly removing irrelevant fine-grained emotion labels based on polarity. No-gating refers to predicting fine-grained emotion categories directly without using polarity. As shown in Table 3, without emotion polarity gating, No-gating mechanism is easy to mislead by local visual noise, making it difficult to directly pinpoint subtle differences for the recognition of fine-grained emotion categories. Besides, hard-gating mechanism is prone to misclassifying the emotion polarity of samples with ambiguous emotions, thereby directly discarding the correct answer and causing irreversible errors. Compared to them, our soft-gating mechanism employs cascaded score products instead of hard dropout, thereby mitigating error accumulation and maintaining a smooth probability distribution for CDVP module.

\begin{wrapfigure}{r}{0.55\textwidth} % r 表示居右，占 0.5 宽度
  \centering
        \setlength{\aboverulesep}{0pt}
\setlength{\belowrulesep}{0pt}
  \vspace{0pt} % 根据需要调整表格上方的间距
  \caption*{Table 4: Discussion for strategies of attention scores.}
  \label{map}
  \small
\begin{tabular}{l|ccc}
\toprule
\rowcolor{gray!40}{Setting}  &  {Emotion6}   & {WebEmo25} & {Abstract8}  \\  
\midrule
Global Average &   \underline{58.59}     &  \underline{24.55} & \underline{32.46}  \\
\rowcolor{gray!15}First-3 Average &   57.58    & 24.30  &  30.26    \\
Last Layer &   {58.42}   &  {24.50} &   {31.58}  \\
\rowcolor{oursburgundy}Last-3 Average& \textbf{59.09}  & \textbf{24.80} & \textbf{33.77} \\
\bottomrule
\end{tabular}
  \vspace{-15pt} % 调整表格下方的间距
\end{wrapfigure}

\textbf{Selection of Attention Map.} In our proposed CDVP module, we calculate the discriminative visual mask by the attention score of MLLMs. Therefore, it is crucial to discuss the selection way of the attention map. We conduct the ablation study under the following four settings: (a) Global Average, which computes the mean attention scores across all Transformer layers, (b) First-3 Average, which considers only the initial three layers, (c) Last Layer, which directly utilizes the attention scores from the last layer, and (d) Last-3 Average, which averages the scores from only the final three layers. As shown in Table 4, we observe that Last-3 Average achieves the best performance. We infer that MLLMs are difficult to establish a semantic understanding of visual images in the early stages of inference. Furthermore, using only the last layer will result in the loss of semantic information and reduce inference stability.

\begin{wrapfigure}{r}{0.6\textwidth} % r 表示居右，占 0.5 宽度
  \centering
        \setlength{\aboverulesep}{0pt}
\setlength{\belowrulesep}{0pt}
  \caption*{Table 5: Discussion for CDVP module.}
  \label{cdvp}
  \small
\begin{tabular}{l|ccc}
\toprule
\rowcolor{gray!40}{Setting}  &  {Emotion6}   & {WebEmo25} & {Abstract8}  \\  
\midrule
\rowcolor{gray!15}Only HERC &   {57.24}     &  {23.15} & {31.58}  \\
 + $S$ &  56.57    &   22.95    &   30.70    \\
\rowcolor{gray!15} + $S^*$ & 57.41     &  23.25     &   32.01    \\
 + $S^*$ +$Q_{ctr}$ &   \underline{58.25}   &  \underline{24.05}     &   \underline{32.89}    \\
\rowcolor{oursburgundy} + $S^*$ +$Q_{ctr}$ + CDP & \textbf{59.09}     &  \textbf{24.80}     &   \textbf{33.77}    \\
\bottomrule
\end{tabular}
\end{wrapfigure}

\textbf{Discussion on CDVP module.} To explore the contributions of dense components in CDVP module, we conduct a progressive ablation study, as shown in Table 5. We use the top-1 emotion predicted by HERC as the baseline. We first observe that re-ranking over the full emotion set \(S\) even causes slight degradation, suggesting that the gain of CDVP does not stem from repeated calls to MLLMs. Besides, we observe that replacing the entire emotion set as the ambiguous set \(S^*\) only improves performance slightly, which indicates that narrowing the range of choices is not the crucial reason for improved performance. Instead, we observe that introducing the contrastive prompt \(Q_{\mathrm{ctr}}\) and contrastive discriminative pruning CDP both significantly improve the performance, which validates our two motivations: 1) Compared to direct selections, contrastive questions are more effective at guiding the model to identify differences and provide correct answers. 2) Forcing the model to focus on the most discriminative visual regions significantly improves the ability to decode affective nuances.

\begin{wrapfigure}{r}{0.55\textwidth} % r 表示居右，占 0.5 宽度
  \centering
      \setlength{\aboverulesep}{0pt}
\setlength{\belowrulesep}{0pt}
  \vspace{-10pt} % 根据需要调整表格上方的间距
  \caption*{Table 6: Results of different pruning strategies.}
  \label{pruning}
  \small
\begin{tabular}{l|ccc}
\toprule
\rowcolor{gray!40}{Dataset}  &  {Emotion6}   & {WebEmo25} & {Abstract8}  \\  
\midrule
Random &  55.56      & 23.70  & 27.63  \\
\rowcolor{gray!15}Query-related &  56.57     & 23.95  & 29.82     \\
FoE-related &  \underline{57.91}    & \underline{24.25}  & \underline{31.58}    \\
\rowcolor{oursburgundy}Ours& \textbf{59.09}  & \textbf{24.80} & \textbf{33.77} \\
\bottomrule
\end{tabular}
  \vspace{-10pt} % 调整表格下方的间距
\end{wrapfigure}

\textbf{Pruning Strategy Discussion.} To demonstrate the effectiveness of contrastive discriminative pruning, we make a comparison of four different pruning strategies. (1) Random Pruning, which prunes a fixed number of tokens at random; (2) Query-related Pruning, which prunes tokens based on their attention scores relative to the overall prompt; (3) FoE-related Pruning proposed in SEPM~\citep{fang2025catch}, which utilizes attention scores derived from specific Focus-on-Emotion prompt to guide the pruning process; and (4) our proposed Contrastive-related Pruning, which identifies critical visual evidence by leveraging attention signals from contrastive prompts to determine the pruned tokens. As shown in Table 6, our contrastive-related pruning strategy achieves the best performance across all datasets, as it could pinpoint the most discriminative visual regions for ambiguous emotions. In contrast, random pruning may lead to the inaccurate loss of emotion-related tokens. Query-related pruning may introduce visual redundancies associated with emotion-irrelevant prompts. While FoE-related pruning successfully isolates broad emotion-related regions, it fails to differentiate between the subtle nuances of semantically similar emotions. These results demonstrate that the coarse-grained emotion-oriented pruning is insufficient. Instead, the fine-grained and discriminative pruning is essential to empower MLLMs to perceive subtle affective nuances.

\begin{wrapfigure}{r}{0.61\textwidth} 
    \centering
    % 如果图片上方间距太大，可以取消注释下面这行进行微调
     \vspace{-10pt} 
    
    % 插入你的图片，确保图片的宽度略小于 wrapfigure 的宽度，留出一点边距
    \includegraphics[width=0.6\textwidth]{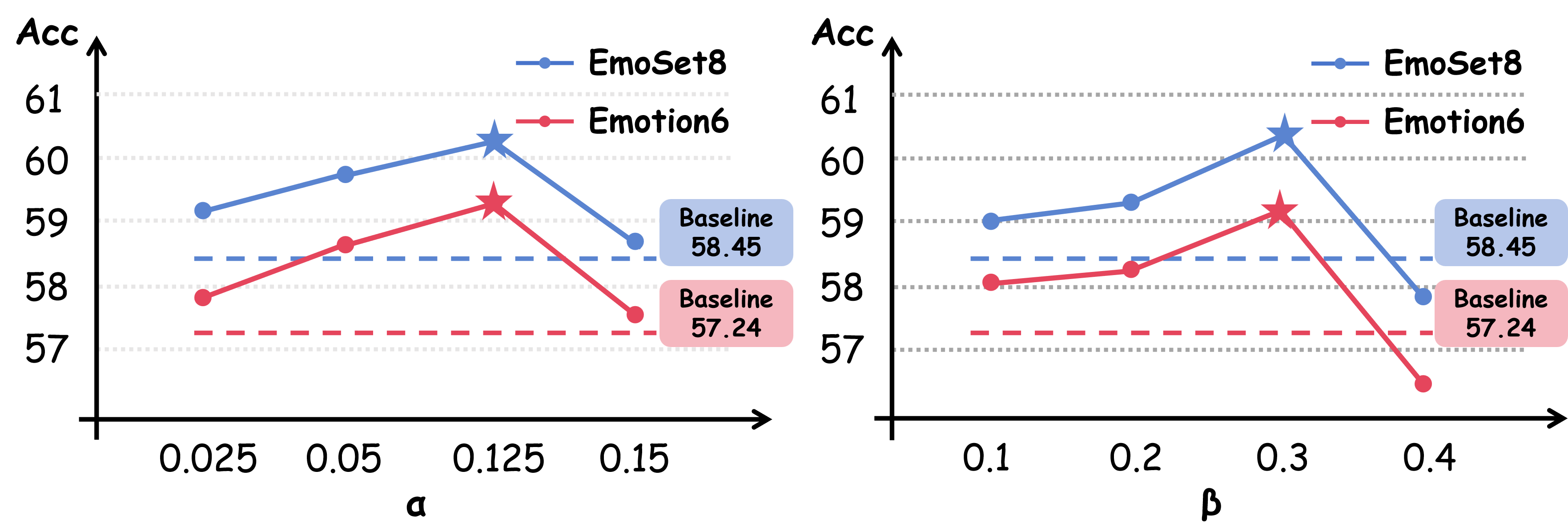} 
    
    % 图片的标题和标签
    \caption*{Figure 3: Parameter sensitivity analysis.}
    \label{conf}
    
    % 如果图片下方间距太大，可以取消注释下面这行进行微调
     \vspace{-10pt}
\end{wrapfigure} 

\textbf{Parameter Sensitivity Discussion.} We conduct a parameter sensitivity discussion for $\alpha$ and $\beta$. As shown in Fig. 3, we first observe that with increasing $\alpha$, the performance initially improves and then decreases. When $\alpha$ falls within an appropriate range, the CDVP module successfully corrects certain misclassified GT emotions. Conversely, when $\alpha$ is excessively high, emotions with very low confidence, which are typically regarded as emotional noise, are included in the consideration, thereby interfering with accurate emotion classification. Furthermore, with increasing $\beta$, the performance initially improves and then declines. This suggests that while a small pruning rate effectively eliminates emotion-irrelevant visual redundancy, a too-high pruning rate leads to the erroneous removal of critical visual cues, thereby compromising emotion accuracy. Based on these observations, we select $\alpha = 0.125$ and $\beta = 0.3$ as the optimal setting to strike a balance between performance and inference efficiency.

% \begin{figure*}[t]        
% \center{\includegraphics[width=1.0\linewidth] {ICLR_FIG_4.png}} 
% \caption*{Figure 4: }{Case study of our proposed DAN framework.}
% \label{fig4}
% \end{figure*}

\subsection{Does DAN alleviate insufficient attribution and discrimination?}\label{4.4} 

  \begin{wrapfigure}{r}{0.5\textwidth} 
     \centering
     \vspace{-20pt}
     \includegraphics[width=0.48\textwidth]{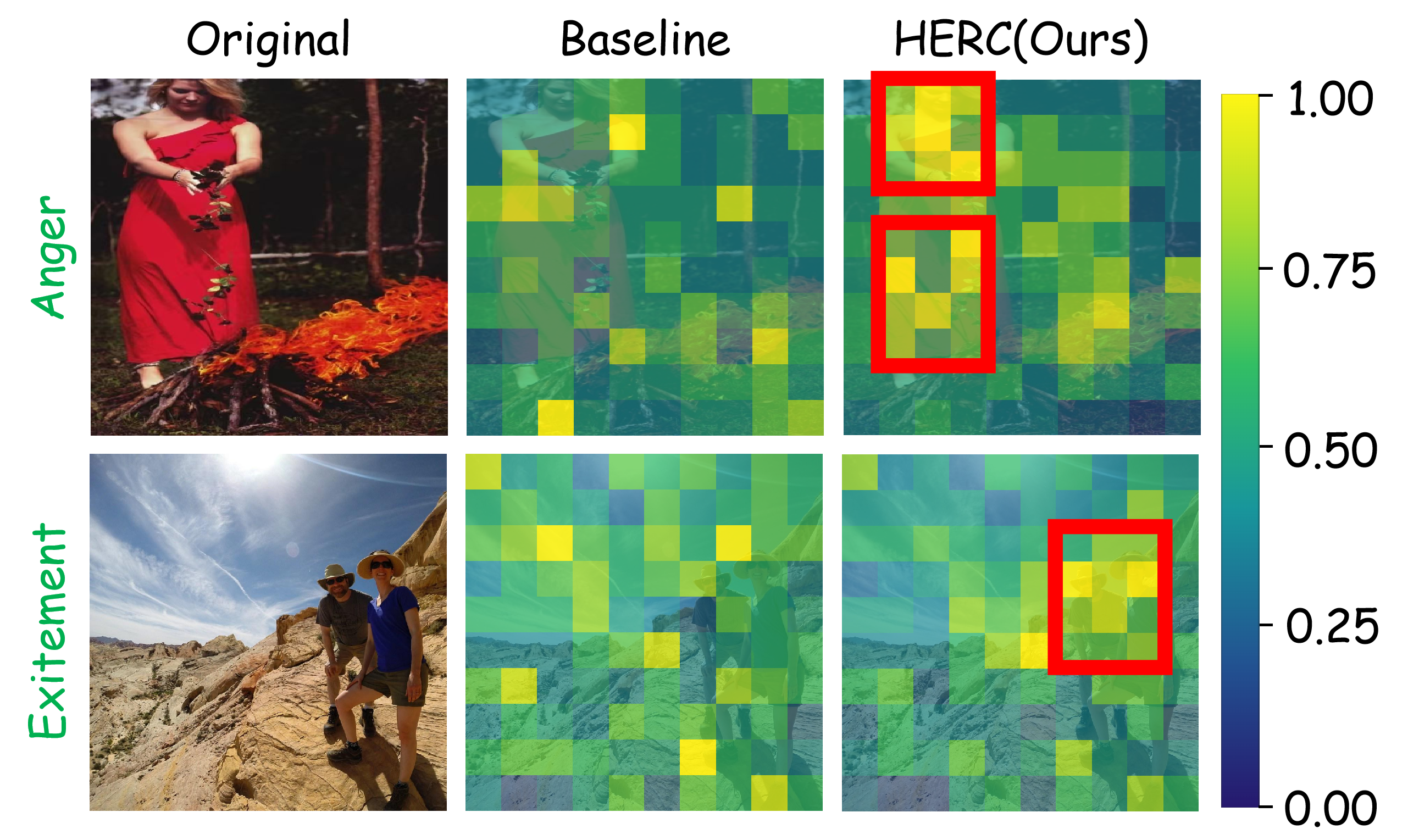} 
     \caption*{Figure 4: Case study of HERC.}
     \label{FIG4}
      \vspace{-15pt}
 \end{wrapfigure}
\textbf{Visualization of Sufficient Attribution.} To verify that our HERC module mitigates insufficient attribution, we extract the attention maps output by HERC module. As shown in Fig. 4, we observe that the baseline only performs global reasoning on these two images, resulting in a uniform attention distribution across the entire image. In contrast, our HERC module successfully focuses attention on key emotion-related regions by guiding the MLLM to mine fine-grained scene and object cues. For example, in the first image, the flames, the woman's facial expression, and the action of burning leaves are highlighted. For the second image, the couple's excited facial expressions are also captured. These results demonstrate that our model successfully mitigates the insufficient attribution associated with global reasoning, thereby effectively focusing attention on fine-grained visual cues.
 
  \begin{wrapfigure}{r}{0.5\textwidth} 
     \centering
    \vspace{-20pt}
     \includegraphics[width=0.48\textwidth]{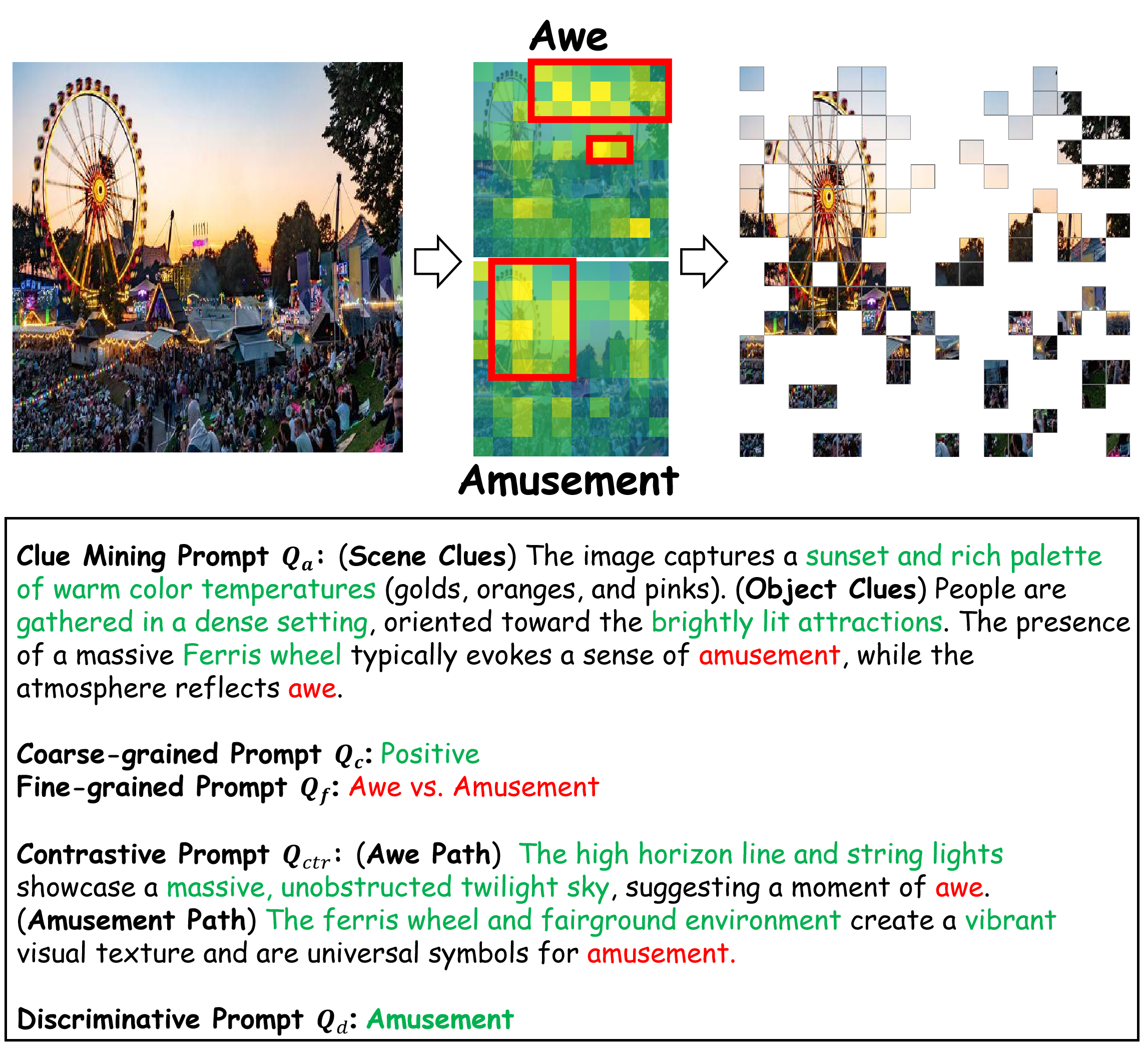} 
     \caption*{Figure 5: Case study of CDVP.}
     \label{FIG4}
      \vspace{-20pt}
 \end{wrapfigure}

\textbf{Visualization of Sufficient Discrimination.} To verify that our CDVP module mitigates insufficient discrimination, we display the reasoning text and pruning process shown in Fig. 5. We observe that DAN selects two ambiguous emotions `awe' and `amusement'. Then we employ the contrastive prompt on these two emotions for two contrastive attention maps. We find that `awe' mainly focuses on the sky and light while `amusement' focuses more on the ferris wheel and playground. Based on this, CDVP performs the discriminative discrepancy calculation to generate the discrepancy map and finally select the correct emotion `amusement' based on it, which demonstrate that CDVP could enhance the discrimination between semantically similar emotions.

% To provide a visualization for our DAN framework, we show a representative case study. As shown in Fig. 4, our proposed HERC module initially extracts hierarchical scene-level (\emph{i.e.,} \textit{`brightly lit attractions'}) and object-level (\emph{i.e.,} \textit{`people are gathered in a dense setting'}) affective clues. Subsequently, the MLLM performs multi-granularity reasoning to determine the affective polarity and select the ambiguous emotion set. Building upon this, the CDVP module computes a differential attention map via contrastive prompting, which yields the corresponding pruned image shown at the top of Fig. 5. Finally, we leverage these discriminative tokens to achieve precise emotional judgment.}

% \textbf{$\ast \ast \ast$ Related works, details of compared baselines, analysis of confidence variance statistics, additional experimental analysis, and details of prompt templates are illustrated in the Appendix $\ast \ast \ast$}

%% file: Sec/conclu.tex
\section{conclusion}

In this paper, we focus on enhancing the emotion understanding of Multimodal Large Language Models (MLLMs) during the inference phase, particularly for ambiguous scenarios with semantically proximal emotions. We propose the Decoding Affective Nuances (DAN) framework, which is composed of two crucial modules: the Hierarchical Emotion Reasoning Chain (HERC) and Contrastive Discriminative Visual Pruning (CDVP).
Specifically, HERC guides the MLLM to a hierarchical emotion reasoning pattern, integrating coarse-to-fine clue mining and soft-gated emotion reasoning. Besides, CDVP enhances the discriminative ability of MLLM by focusing on ambiguous candidate emotions. We first design a conditional algorithm to select ambiguous emotions, and leverage a contrastive prompt to constrain the MLLM focus on discriminative visual regions associated with candidate emotions, followed by differential visual pruning to generate a visual dropping mask. Finally, a discriminative emotion re-selection is executed based on the dropping mask. Our approach enhances the discriminative capability and achieves significant accuracy improvements in emotion recognition without additional training or annotation. Extensive experiments on several benchmarks demonstrate the effectiveness and rationality of the framework and components.

%% file: Sec/appendix.tex
\subsection{related work}
\textbf{Emotion Recognition in MLLMs.}
The emergence of Multimodal Large Language Models has ushered in a new paradigm in the field of multimodal perception and understanding tasks, characterized by their excellent reasoning and generative capabilities. By integrating powerful MLLMs with sophisticated encoder frameworks, these models have demonstrated remarkable proficiency on a variety of tasks across different modalities~\citep{dai2023instructblip,luofeast,cai2024vip,chen2024sharegpt4v,hong2026emostyle}. Despite their prowess in general tasks, applying MLLMs to emotion-related tasks presents significant challenges. Emotion recognition is inherently subjective and usually fraught with ambiguity, requiring to disentangle fine-grained emotion nuances. Current MLLMs usually struggle to distinguish between emotionally proximal categories and become distracted by redundant visual information. To mitigate these issues, recent research explored various strategies to leverage MLLMs for emotion-related tasks. Some works ~\citep{lian2026merbench,wang2025emotion,lian2025ov,wang2023improving} construct large-scale affective datasets and focus on supervised instruction tuning to enhance emotion understanding. AffectGPT ~\citep{lian2025affectgpt} establishes a descriptive emotion dataset with 2K fine-grained emotion categories and designs a pre-fusion operation to enhance multimodal integration. Emotion-LLaMA ~\citep{cheng2024emotion} introduces an emotion-specific encoder to seamlessly integrate multimodal inputs and aligns multimodal features with instruction tuning, thereby enhancing the understanding and reasoning capabilities about emotions. However, these methods require carefully designed datasets and fine-tuning, leading to high manual and training costs. In this paper, we aim to enhance the emotion understanding capabilities of MLLMs during inference without additional annotations and training costs to achieve more efficient emotion understanding.

\textbf{Training-Free Methods.}
Despite impressive progress, MLLM-based methods with fine-tuning remain limited by extremely high annotation and computational costs. Therefore, training-free methods, which leverage the inherent capabilities of frozen pre-trained models during the inference phase, have garnered significant attention from the research community ~\citep{tang2025reason,wangvideorft,wang2025embracing}. Typical training-free methods involve techniques such as prompt engineering to optimize the textual instructions to guide the model to perform more complex reasoning, in-context strategies to leverage the model’s in-context learning abilities, token pruning to dynamically re-weight tokens to ensure the model focuses on the most informative regions, and multi-agent collaboration to assign different roles to multiple models to refine the output through iterative dialogue and self-reflection. MM-PEAR-CoT ~\citep{li2025multimodal} design a PEAR CoT prompt based on preliminaries, question, answer, and reason, to generate text-based reasoning processes and zero-shot sentiment prediction results. EmoGist ~\citep{seoh2025emogist} proposes an in-context emotion learning strategy, which re-generates multiple descriptions of emotion labels by analyzing the clusters of example images belonging to each label and retrieves a version of description based on the cosine similarity of images to cluster centroids for classification at test time. SEPM ~\citep{fang2025catch}, which performs visual token pruning by directing the attention of coarse-grained emotion prediction to relevant emotional cues in images. MERMAID ~\citep{yang2025mermaid}designs a multi-agent framework to address the ambiguity of emotion recognition in the wild, which consists of a multi-perspective reflection agent, an emotion-guided augmentation agent, and a cross-modal verification agent.

\vspace{20pt}

\subsection{Experimental Setup}
\textbf{Dataset.} We evaluate the performance of our framework on four public emotion recognition benchmarks, EmoSet~\citep{yang2023emoset}, WebEmo~\citep{panda2018contemplating}, Emotion6~\citep{peng2015mixed}, and Abstract~\citep{machajdik2010affective}. The EmoSet dataset contains $\sim$118k images from social media and artworks. They are annotated by both machines and humans with 8 basic emotions. The WebEmo dataset contains $\sim$2k images collected from the web and includes emotion labels at two granularity levels with 7 and 25 categories. The Emotion6 dataset contains 1980 images and are annotated with 6 basic emotions. The Abstract dataset contains 228 abstract paintings, including colors and textures. They are labeled with 8 basic emotions through manual voting. Besides, we leverage emotion accuracy (Acc) as the evaluation metric for all experiments.

\textbf{Implementation Details.} We leverage four MLLMs to evaluate the effectiveness of our framework, including Qwen2.5-VL-7B-Instruct~\citep{qwen2025qwen25technicalreport}, Qwen3-VL-4B-Instruct~\citep{bai2025qwen3}, Qwen3-VL-8B-Instruct~\citep{bai2025qwen3}, and InternVL3.5-8B~\citep{wang2025internvl3}. We set the variance threshold $\alpha=0.125$, the dropping ratio $\beta=0.3$, and the truncation threshold $k=2$. Since our framework is inference-time and training-free, all experiments are only conducted on a single NVIDIA A800 GPU with 80GB of memory. Unless otherwise specified, all ablation studies are conducted on the Qwen3-VL-8B-Instruct model, which has the largest number of parameters.

\subsection{Details of Baselines}
We introduce four baseline methods in our experiments:

%To comprehensively evaluate the performance of the proposed DAN framework, we select several state-of-the-art (SOTA) methods as baselines. First, the Zero-shot approach is employed as a fundamental baseline to characterize the inherent emotion understanding capabilities of MLLMs. Furthermore, we include Zero-shot-CoT~\citep{kojima2022large} to benchmark our proposed HERC against standard Chain-of-Thought paradigms. Finally, we compare our method with two representative pruning-based methods, namely SparseVLM~\citep{zhang2024sparsevlm} and SEPM~\citep{fang2025catch}, to highlight the superiority of our proposed CDVP in capturing discriminative visual tokens.
%For the Zero-shot-CoT method, the prompt is designed as ``\textit{Let’s think step by step, then answer with the option’s letter from the given choices directly.}""

\textbf{Zero-shot.} Firstly, we leverage the zero-shot configuration as a fundamental baseline to characterize the inherent emotion understanding capabilities of MLLMs. Specifically, the prompt is directly set to ``\textit{Which of the following descriptions best represents the image?} \verb|[Emotion Set]| \textit{Answer directly with the number of the chosen option.}``.

\textbf{Zero-shot CoT.} Besides, we introduce the Zero-shot-CoT configuration to make a comparison with the native reasoning ability of MLLMs. Specifically, the prompt is directly set to ``\textit{Let's think step by step. Which of the following descriptions best represents the image?} \verb|[Emotion Set]| \textit{Answer directly with the number of the chosen option.}``.

% \textbf{SparseVLM.} SparseVLM is a training-free framework that prunes irrelevant visual tokens by selecting visual-relevant text tokens to rate the significance of vision tokens within the self-attention matrix. Although it achieves visual pruning, it aligns only based on textual semantics and fails to focus on visual regions specific to emotional semantics. Therefore, it serves as a foundation for pruning-based approaches.

\textbf{SEPM.} SEPM is a training-free framework to sharpen emotion perception via emotion-related visual pruning. Despite achieving credible progress, SEPM indiscriminately focuses on all emotion-related regions, which weakens its discriminative ability for semantically similar emotion categories, thereby hindering its ability to handle challenging samples with ambiguity and vagueness.

\subsection{Why we choose confidence variance as a standard?}\label{5}

\begin{wrapfigure}{r}{0.5\textwidth} 
    \centering
    % 如果图片上方间距太大，可以取消注释下面这行进行微调
     \vspace{-10pt} 
    
    % 插入你的图片，确保图片的宽度略小于 wrapfigure 的宽度，留出一点边距
    \includegraphics[width=0.48\textwidth]{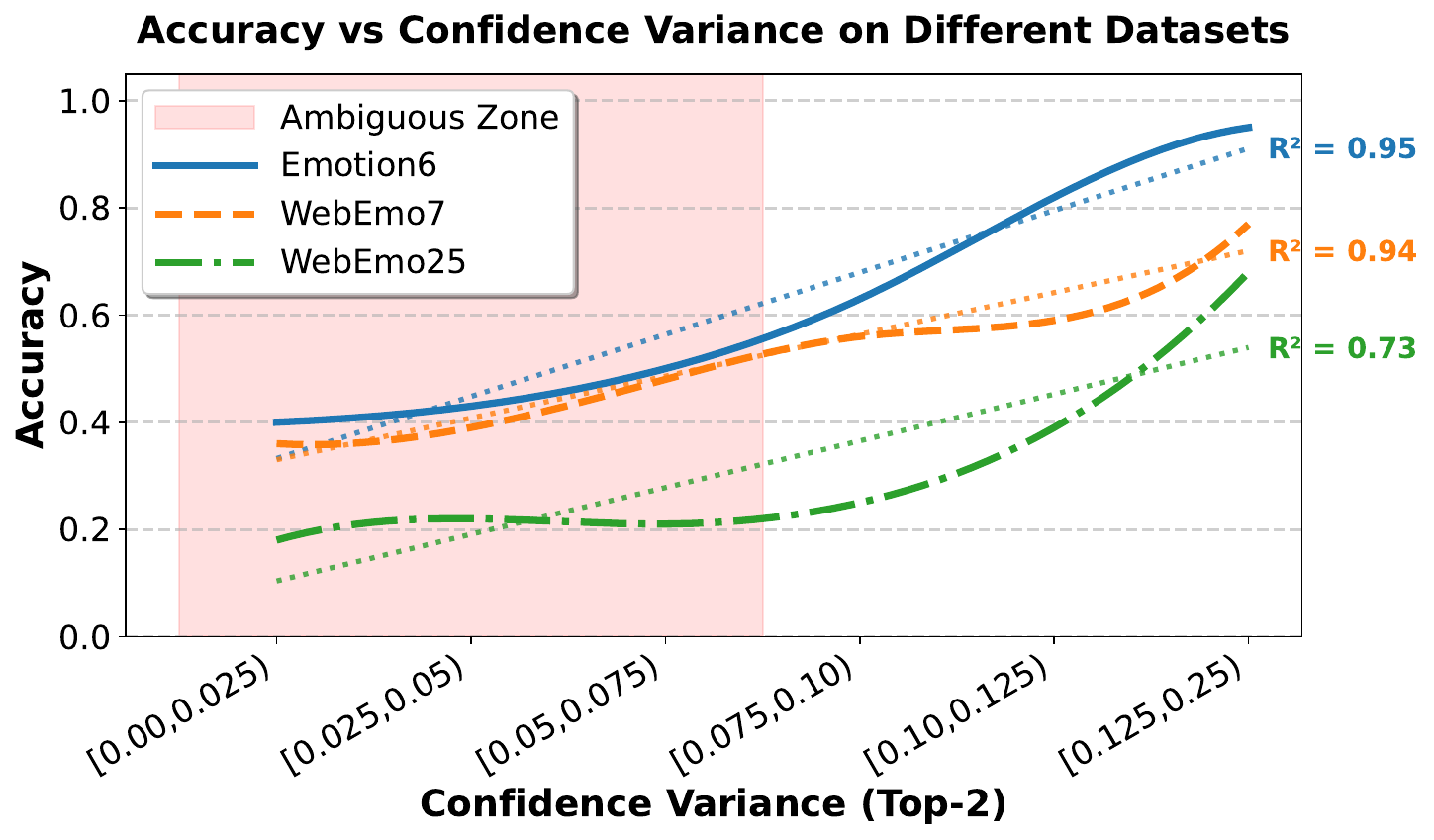} 
    
    % 图片的标题和标签
    \caption*{Figure A.1: Visualization of variance distribution.}
    \label{conf}
    
    % 如果图片下方间距太大，可以取消注释下面这行进行微调
     \vspace{-10pt}
\end{wrapfigure}

\textbf{Is Confidence Variance a Gold Standard?} To verify the motivation of our proposed CDVP module, we make a visualization of the variance distribution. Based on the preliminary emotion distribution generated by our HERC module, we calculate the confidence variance between the top-2 candidate emotions. The statistical distribution is shown in Fig. A.1. We observe that when the variance increases, the emotion accuracy of the corresponding subsets improves progressively. This trend suggests that the variance could serve as a robust metric to quantify the ambiguity of MLLM in emotion judgment. Specifically, a lower variance indicates that MLLM struggles to differentiate between semantically proximal emotions, leading to a higher possibility for error. These findings not only validate the motivation behind our CDVP module but also verify the rationality of employing variance as a gating metric to determine the necessity of performing the CDVP module.

\begin{wrapfigure}{r}{0.58\textwidth} 
  \centering
            \setlength{\aboverulesep}{0pt}
\setlength{\belowrulesep}{0pt}
  \vspace{-10pt}
  \caption*{Table A.1: Result for different truncation threshold $k$.}
  \label{prompt}
  \small
\begin{tabular}{l|cccc}
\toprule
\rowcolor{gray!40}{$k$}  &  {Emotion6}   & {EmoSet} & {WebEmo7} & {WebEmo25}  \\  
\midrule
\rowcolor{gray!15}1 & 72.39	&72.36	&76.70  &29.39 \\
2 & \underline{91.41}	&\underline{94.86}	&\underline{92.10}	&\underline{41.22}   \\
\rowcolor{oursburgundy}3 & \textbf{91.41}	&\textbf{94.94}	&\textbf{93.00}	&\textbf{42.11} \\
\bottomrule
\end{tabular}
  \vspace{-10pt} 
\end{wrapfigure}

\textbf{Is $k=2$ enough for discrimination?} In our proposed CDVP module, we introduce a truncation threshold $k=2$ to constrain the number of the ambiguous emotion set, thereby seeking a balance between classification accuracy and computational efficiency. A natural question arises: Is $k=2$ sufficient to bolster the discriminative capacity? To answer this, we conduct a statistical analysis on the preliminary emotion distribution from HERC. {Specifically, we evaluate the $Recall@k$ of the ambiguous set to the ground-truth at $k \in \{1, 2, 3\}$, that is, checking whether the ground truth is included in the top-$k+1$ predictions. It ensures that our choice of $k=2$ does not prematurely exclude the ground-truth and lead to an irreversible failure.} As shown in Table A.1, we observe that for benchmarks except WebEmo25, most samples have their ground-truth labels successfully covered when $k=2$. Scaling from $k=1$ to $k=2$ yields a substantial improvement on $Recall@k$. Conversely, scaling from $k=2$ to $k=3$ brings negligible gains. Consequently, jointly considering precision and efficiency, we set the threshold at $k=2$. {Additionally, it is worth noting that the $Recall@k$ on WebEmo25 under all settings is extremely low. Even for $k=3$, the $Recall@k$ is still only 42.11\%. We infer that WebEmo25 contains 25 fine-grained emotional categories, which greatly increases the difficulty of human labeling. Therefore, it significantly exacerbates the subjective bias and ambiguity of human annotation, thereby objectively lowering the upper limit of model prediction performance. Meanwhile, constructing high-quality emotional labels in complex scenarios and designing robust discriminative models based on them constitute an important direction for our future research.}

\begin{wrapfigure}{r}{0.55\textwidth} 
    \centering
    % 如果图片上方间距太大，可以取消注释下面这行进行微调
     \vspace{-10pt} 
    % 插入你的图片，确保图片的宽度略小于 wrapfigure 的宽度，留出一点边距
    \includegraphics[width=0.5\textwidth]{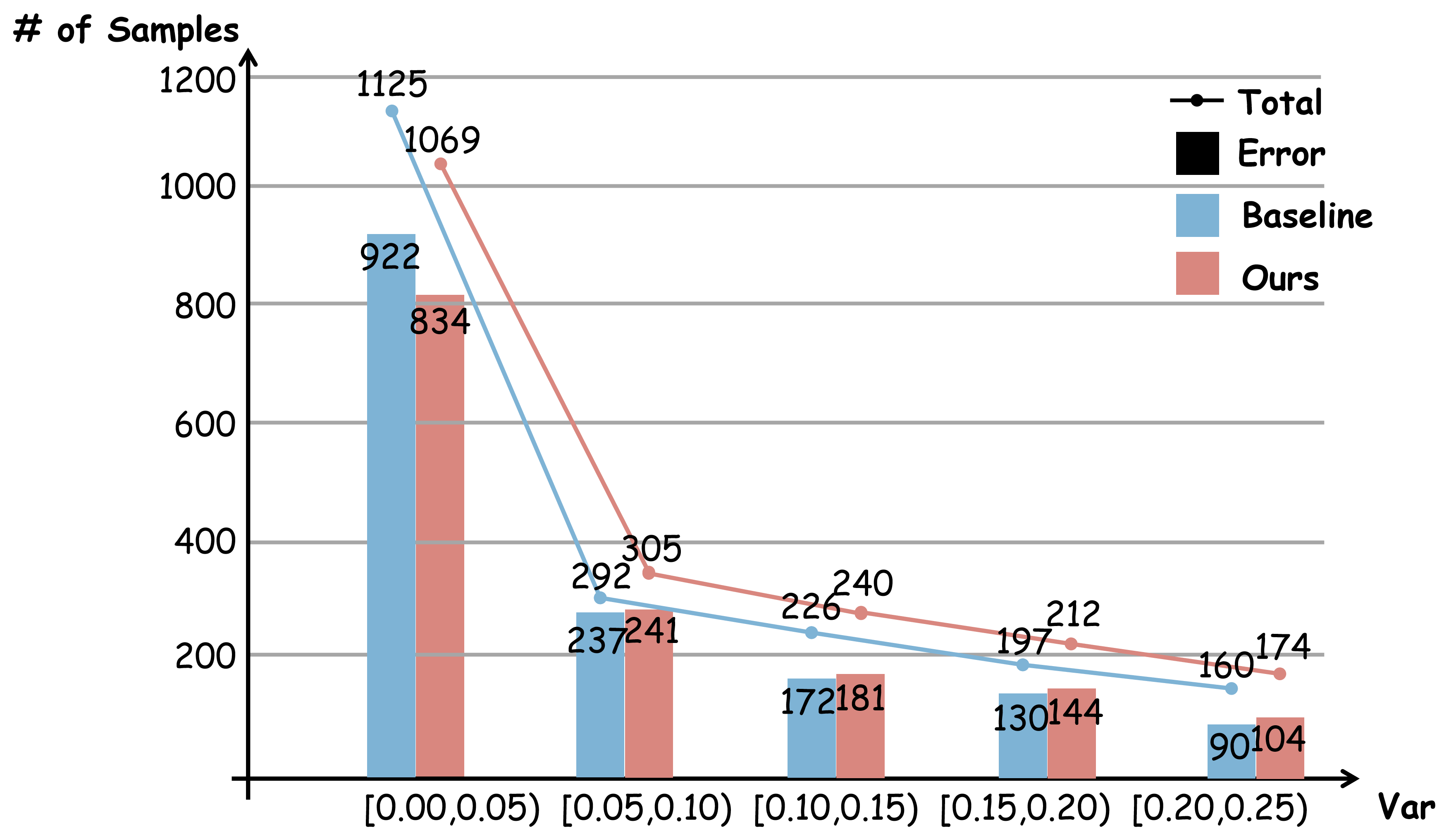} 
    
    % 图片的标题和标签
    \caption*{Figure A.2: Comparison of discriminative abilities.}
    \label{conf}
    % 如果图片下方间距太大，可以取消注释下面这行进行微调
     \vspace{-10pt}
\end{wrapfigure}

\textbf{Comparison of Discrimination.} To verify whether DAN framework substantially enhances the discrimination capacity of MLLMs, we perform a statistical analysis across different top-2 confidence variance intervals on WebEmo25 and compare the number of error samples against the baseline model. As shown in Fig. A.2, our framework significantly decreases the number in low-variance intervals that represent high semantic ambiguity. We infer that this substantial improvement is fundamentally empowered by the proposed CDVP module. Specifically, CDVP accurately isolates the ambiguous emotions and leverages contrastive prompting to synthesize a dropping mask, forcing MLLMs to anchor on the most discriminative visual regions. Consequently, the MLLMs execute a precise affective re-selection based on these discriminative visual regions, mitigating the risk of misclassifications.

\subsection{Why are tokens that exhibit significant differences between attention maps of two emotions selected for pruning?}\label{question2}

Previous methods such as SEPM use a “Focus on emotion” prompt to identify visual tokens that are generally relevant to emotional perception, and prune tokens with low emotion relevance. While effective for removing emotion-irrelevant visual redundancy, such a relevance-based criterion does not explicitly determine which regions distinguish two semantically similar candidate emotions.

CDVP extends this idea by constructing candidate-conditioned contrastive prompts. For each ambiguous candidate emotion, the corresponding attention map measures how strongly each visual token is associated with that candidate under comparison with its competitors. A token receiving similarly high attention for both candidates may be emotion-related, but it provides shared evidence and therefore has limited discriminative value. For example, a human face may be important for recognizing both sadness and disappointment. In contrast, a token whose attention differs substantially between the two candidate-conditioned maps is associated asymmetrically with the competing emotions. Such asymmetric relevance provides more informative evidence for deciding which candidate better explains the image.

Therefore, we use the attention discrepancy as a proxy for inter-emotion discriminativeness. Specifically, CDVP assigns higher scores to tokens with larger attention discrepancies, retains these discriminative tokens, and prunes low-discrepancy tokens that mainly encode shared or irrelevant information. We agree that attention discrepancy alone does not constitute a theoretical guarantee that a region is causally decisive. We therefore interpret it as a candidate-conditioned discriminativeness score rather than direct causal evidence. Its effectiveness is supported empirically by our ablations: Table 5 shows an additional improvement when contrastive discriminative pruning is added on top of the ambiguous candidate set and contrastive prompting, while Table 6 shows that our discrepancy-based pruning consistently outperforms random, query-related, and FoE-related pruning. These comparisons indicate that the gain comes from preserving candidate-discriminative visual information rather than from generic token removal or emotion-related attention alone.

\subsection{Additional Experimental Analysis}

\begin{wrapfigure}{r}{0.55\textwidth} % r 表示居右，占 0.5 宽度
  \centering
    \setlength{\aboverulesep}{0pt}
\setlength{\belowrulesep}{0pt}
  \vspace{-10pt} % 根据需要调整表格上方的间距
  \caption*{Table A.2: Results of emotion polarity classification.}
  \label{polar}
  \small
\begin{tabular}{l|ccc}
\toprule
\rowcolor{gray!40}{Dataset}  &  {Emotion6}  & EmoSet8   & {Abstract8}  \\ 
\midrule
SEPM &     82.83    &  90.13      & 69.74     \\
\rowcolor{gray!15}Scene &    \underline{87.37}    &  \underline{95.17}      & \underline{76.75}      \\
Object &   86.03    &  92.91      & 71.93      \\
\rowcolor{oursburgundy}HERC(Ours)&  \textbf{89.90}  &  \textbf{96.44}   & \textbf{80.26} \\
\bottomrule
\end{tabular}
  \vspace{-10pt} % 调整表格下方的间距
\end{wrapfigure}

\textbf{Scene/Object Clue Discussion.} Within our HERC module, the fine-grained emotion set is strictly dependent on the outcomes of the coarse-grained stage. Consequently, achieving a stable and accurate coarse-grained polarity recognition serves as the cornerstone for enhancing overall framework performance. We conduct an ablation study to investigate the influence of scene/object-level cues on coarse-grained polarity determination. As shown in Table A.2, we observe that scene-level cues play a more crucial role in polarity assessment. Our analysis suggests that an overemphasis on localized object details may trigger visual hallucinations in MLLMs, leading to mistakes for polarity judgments. Conversely, scene-level cues characterize the emotion polarity from a holistic perspective, focusing on the global environment and atmosphere, which proves to be more stable and reliable, while effectively mitigating the risk of hallucinations.

\begin{wrapfigure}{r}{0.55\textwidth} 
  \centering
          \setlength{\aboverulesep}{0pt}
\setlength{\belowrulesep}{0pt}
  \vspace{-10pt}
  \caption*{Table A.3: Discussion for prompt injection strategies.}
  \label{prompt}
  \small
\begin{tabular}{l|ccc}
\toprule
\rowcolor{gray!40}{Setting}  &  {Emotion6}   & {WebEmo25} & {Abstract8}  \\  
\midrule
\rowcolor{gray!15}Entire &    58.59    & 24.60  & 32.01  \\
Module-based &  \underline{58.75}  & \underline{24.75}   &  \underline{33.33}    \\
\rowcolor{oursburgundy}Alone& \textbf{59.09}  & \textbf{24.80} & \textbf{33.77} \\
\bottomrule
\end{tabular}
  \vspace{-10pt} 
\end{wrapfigure}

\textbf{Prompt Sensitivity Discussion.} In our proposed DAN framework, we employ extensive prompt engineering to activate the emotion understanding capabilities of MLLMs. Consequently, it is necessary to discuss the sensitivity to different prompt injection strategies. Specifically, we design three distinct settings: (a) Entire Injection: Most prompts are aggregated into a single, long-context sequence for one-shot input. Notably, as the discriminative prompt $Q_d$ needs additional visual mask inputs, it remains separate, while all other prompts, including $Q_a, Q_c, Q_f, Q_{ctr}$ are concatenated into one entire prompt. (b) Module-based Injection: Prompts are grouped based on proposed modules. Specifically, $Q_a, Q_c$, and $Q_f$, which constitute the HERC module, are grouped into a single prompt. (c) Alone Injection: Each individual prompt is injected into the MLLM independently and sequentially, maintaining the granularity of each reasoning stage. As shown in Table A.3, we observe that the accuracy fluctuations across the three prompting strategies are remarkably negligible across all datasets. This consistency strongly demonstrates that the efficacy of our proposed method is inherent and based on architectural design, rather than prompt engineering. Furthermore, this also validates the robustness and stability of the overall framework.

\begin{wrapfigure}{r}{0.5\textwidth} % r 表示居右，占 0.5 宽度
  \centering
              \setlength{\aboverulesep}{0pt}
\setlength{\belowrulesep}{0pt}
  \vspace{-10pt} 
  \caption*{Table A.4: Performance on FGIC task.}
  \label{map}
  \small
\begin{tabular}{l|cccc}
\toprule
\rowcolor{gray!40}{Setting}  &  {CUB}   & {Cars} & {Aircraft} & Dogs  \\  
\midrule
\rowcolor{gray!15}MCQA~\citeyearpar{atabuzzaman2025zero} &  23.30      & 21.94  & 31.14 &22.50 \\
FINER~\citeyearpar{kim2024finer} &   20.67   & \textbf{29.97} & 32.29  & \textbf{36.30} \\
\rowcolor{oursburgundy}DAN& \textbf{25.90} & \underline{28.64} &  \textbf{33.97}  &  \underline{34.50}  \\
\bottomrule
\end{tabular}
  \vspace{-10pt} % 调整表格下方的间距
\end{wrapfigure}

\textbf{Performance on FGIC task. }To evaluate the scalability and generalization of the proposed DAN framework, we extend its application to Fine-Grained Image Classification (FGIC) task and benchmark its performance. Extensive experiments are conducted across four public datasets, including CUB-200-2011~\citep{wah2011caltech}, Stanford Cars~\citep{krause20133d}, FGVC-Aircraft~\citep{maji2013fine}, and Stanford Dogs~\citep{dataset2011novel}. We compare our framework with two SOTA methods, including MCQA~\citep{atabuzzaman2025zero} and FINER~\citep{kim2024finer}. As shown in Table A.4, our framework achieves highly competitive performance, surpassing existing methods on the CUB-200-2011 and FGVC-Aircraft benchmarks, while performing on par with the best model on the Stanford Cars and Stanford Dogs datasets. The result demonstrates that our framework is not limited to emotion recognition but is effectively scalable to broader tasks requiring fine-grained recognition. This success is fundamentally attributed to the proposed CDVP module, which leverages contrastive prompting to enhance the discriminative sensitivity to subtle visual nuances.

\begin{table*}[h]
\centering
\setlength{\aboverulesep}{0pt}
\setlength{\belowrulesep}{0pt}
\caption*{Table A.5: Result for the trigger rate of CDVP.}
\scalebox{1.0}{
\begin{tabular}{l|ccccc}
\toprule
\rowcolor{gray!40}{Setting}  &  {Emotion6} & EmoSet8 & WebEmo7  & {WebEmo25} & {Abstract8}  \\  
\midrule
Qwen2.5-VL-7B-Instruct &  41.25   &  46.01  & 56.50  & 67.70  &  61.40 \\
\rowcolor{gray!15}Qwen3-VL-4B-Instruct &   59.26  &62.44   & 69.85  & 78.20  & 71.93  \\
Qwen3-VL-8B-Instruct&   38.22  & 43.33  &  50.30 & 58.45  &  53.51 \\
\rowcolor{oursburgundy}InternVL3.5-8B & 40.57    & 44.68  &  52.15 & 60.05  & 50.88  \\
\bottomrule
\end{tabular}}
\end{table*}

\textbf{Trigger Rate for CDVP.} To intuitively demonstrate the role of our proposed CDVP, we statistically analyze the trigger rate of the CDVP module in each dataset. As shown in Table A.5, we observe a significant decrease in the trigger rate of the CDVP module as the model parameter size increases. We attribute this to the enhanced reasoning capabilities of larger models, which tend to be more confident in their preliminary judgments. Consequently, the confidence variance between top candidate emotions increases, naturally reducing the frequency of triggering the CDVP module. Furthermore, across different datasets, the trigger rates on WebEmo25 is notably higher than those on Emotion6, EmoSet8, and WebEmo7. This suggests that a richer set of fine-grained categories inherently escalates the difficulty of emotional reasoning, causing the model to generate more ambiguous and hesitant predictions. However, despite containing only 8 emotion categories, the Abstract8 dataset also exhibits a substantially high CDVP trigger rate. We infer that because this dataset consists of abstract art paintings, extracting implicit emotional semantics from them is fundamentally more challenging than analyzing real-life human scenarios. This intrinsic visual complexity drives up the ambiguity of the model's predictions, thereby activating the CDVP module more frequently.

\begin{wrapfigure}{r}{0.5\textwidth} 
    \centering
    % 如果图片上方间距太大，可以取消注释下面这行进行微调
     \vspace{-20pt} 
    
    % 插入你的图片，确保图片的宽度略小于 wrapfigure 的宽度，留出一点边距
    \includegraphics[width=0.48\textwidth]{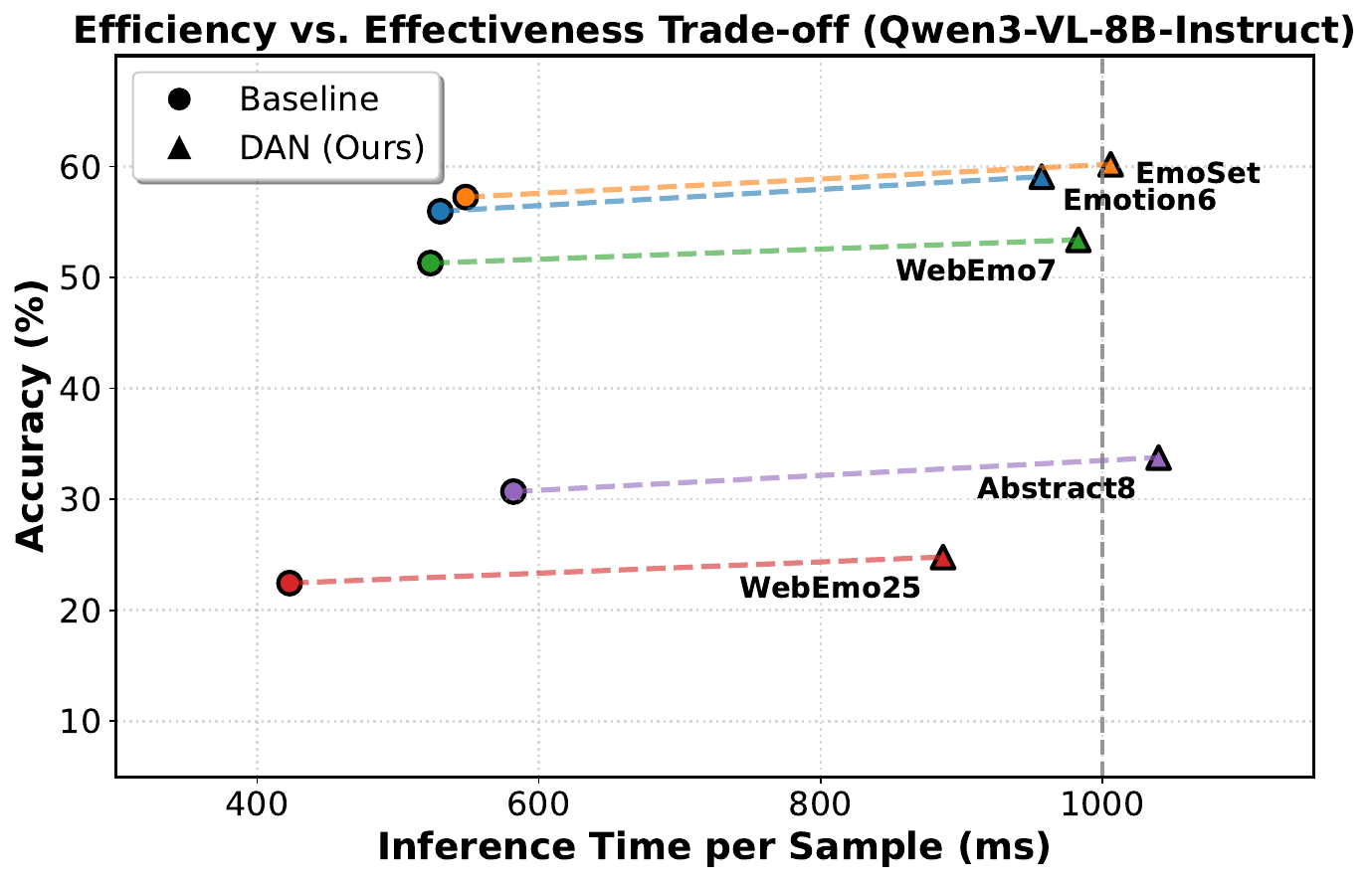} 
    
    % 图片的标题和标签
    \caption*{Figure A.3: Visualization of efficiency analysis.}
    \label{conf}
    
    % 如果图片下方间距太大，可以取消注释下面这行进行微调
     \vspace{0pt}
\end{wrapfigure}
\textbf{Efficiency Analysis.}
Considering that DAN is a training-free and inference-time framework, it is necessary to limit computational overhead to an acceptable range while enhancing performance. Therefore, we conduct an efficiency analysis to evaluate the computational overhead. As shown in Fig. A.3, compared with the baseline method, due to the additional inference of the proposed CDVP module, it did indeed lead to a certain increase in the average inference time per sample. However, it still remains almost within 1 second across all datasets. We consider such efficiency to be user-friendly for practical applications. In the future, we will also explore more efficient and powerful models for emotional reasoning.

\section{Details of Prompt Templates}

\begin{querybox}
    {\textbf{\textcolor{warmtitle}{Clue Mining Prompt $\mathcal{Q}_a$:}}} \\
    \textit{Please identify potential affective-arousing clues in the image following these steps:}\\
    \textit{\textbf{Step 1}: Identify scene-level clues that reflect emotions (e.g., lighting and colors).}\\
    \textit{\textbf{Step 2}: Identify object-level clues that reflect emotions (e.g., facial and body expressions).}\\
    \\\\
    {\textbf{\textcolor{warmtitle}{Coarse-grained Prompt $\mathcal{Q}_c$:}}} \\
    \textit{Based on the above analysis, choose an option that best represents the image:}
    \begin{center}
        \textbf{1}. Positive \qquad \textbf{2}. Negative
    \end{center}
    \textit{Positive emotions include $[{\rm PES}]$. Negative emotions include $[{\rm NES}]$. Answer directly with the number of the chosen option.}
    \\\\
    {\textbf{\textcolor{warmtitle}{Fine-grained Prompt $\mathcal{Q}_f$:}}} \\
    \textit{Based on the above analysis, choose an option that best represents the image:}
    \begin{center}
     \textbf{1}. $e_1$ \qquad \textbf{2}. $e_2$ \qquad $\cdots$ \qquad \textbf{N}. $e_N$
    \end{center}
    \textit{Answer directly with the number of the chosen option.}
    \\\\
    {\textbf{\textcolor{warmtitle}{Contrastive Prompt $\mathcal{Q}_{ctr}(s^*_i,S^*)$:}}} \\
    \textit{Please analyze that why should this image be categorized as {$s^*_i$} rather than {$S^*\setminus s^*_i$}? Please pinpoint the visual cues that support this distinction.}
    \\\\
    {\textbf{\textcolor{warmtitle}{Discriminative Prompt $\mathcal{Q}_{d}$}}} \\
    \textit{Based on the discriminative visual image, choose an option that best represents the image:}
    \begin{center}
        \textbf{1}. $s^*_1$ \qquad \textbf{2}. $s^*_2$ \qquad $\cdots$ \qquad \textbf{N$^*$}. $s^*_{N^*}$
    \end{center}
    \textit{Answer directly with the number of the chosen option.}
\end{querybox}

%% file: iclr2026_conference.bib
@article{dai2023instructblip,
  title={Instructblip: Towards general-purpose vision-language models with instruction tuning},
  author={Dai, Wenliang and Li, Junnan and Li, Dongxu and Tiong, Anthony and Zhao, Junqi and Wang, Weisheng and Li, Boyang and Fung, Pascale N and Hoi, Steven},
  journal={Advances in neural information processing systems},
  volume={36},
  pages={49250--49267},
  year={2023}
}

@inproceedings{luofeast,
  title={Feast Your Eyes: Mixture-of-Resolution Adaptation for Multimodal Large Language Models},
  author={Luo, Gen and Zhou, Yiyi and Zhang, Yuxin and Zheng, Xiawu and Sun, Xiaoshuai and Ji, Rongrong},
  booktitle={The Thirteenth International Conference on Learning Representations}
}

@inproceedings{cai2024vip,
  title={Vip-llava: Making large multimodal models understand arbitrary visual prompts},
  author={Cai, Mu and Liu, Haotian and Mustikovela, Siva Karthik and Meyer, Gregory P and Chai, Yuning and Park, Dennis and Lee, Yong Jae},
  booktitle={Proceedings of the IEEE/CVF Conference on Computer Vision and Pattern Recognition},
  pages={12914--12923},
  year={2024}
}

@inproceedings{chen2024sharegpt4v,
  title={Sharegpt4v: Improving large multi-modal models with better captions},
  author={Chen, Lin and Li, Jinsong and Dong, Xiaoyi and Zhang, Pan and He, Conghui and Wang, Jiaqi and Zhao, Feng and Lin, Dahua},
  booktitle={European Conference on Computer Vision},
  pages={370--387},
  year={2024},
  organization={Springer}
}

@article{cheng2024emotion,
  title={Emotion-llama: Multimodal emotion recognition and reasoning with instruction tuning},
  author={Cheng, Zebang and Cheng, Zhi-Qi and He, Jun-Yan and Sun, Jingdong and Wang, Kai and Lin, Yuxiang and Lian, Zheng and Peng, Xiaojiang and Hauptmann, Alexander G},
  journal={Advances in Neural Information Processing Systems},
  volume={37},
  pages={110805--110853},
  year={2024}
}

@inproceedings{lian2025affectgpt,
  title={AffectGPT: A New Dataset, Model, and Benchmark for Emotion Understanding with Multimodal Large Language Models},
  author={Lian, Zheng and Chen, Haoyu and Chen, Lan and Sun, Haiyang and Sun, Licai and Ren, Yong and Cheng, Zebang and Liu, Bin and Liu, Rui and Peng, Xiaojiang and others},
  booktitle={International Conference on Machine Learning},
  pages={36993--37014},
  year={2025},
  organization={PMLR}
}

@article{lian2026merbench,
  title={Merbench: A unified evaluation benchmark for multimodal emotion recognition},
  author={Lian, Zheng and Sun, Licai and Ren, Yong and Gu, Hao and Sun, Haiyang and Chen, Lan and Liu, Bin and Tao, Jianhua},
  journal={IEEE Transactions on Pattern Analysis and Machine Intelligence},
  year={2026},
  publisher={IEEE}
}

@inproceedings{wang2025emotion,
  title={Emotion-Qwen-VL: A fully fine-tuned multimodal large language model for micro-expression visual question answering},
  author={Wang, Yujing and Fang, Ruotong and Huang, Xing and Han, Zhiyuan and Lin, Xiaoqing and Shan, Yuhao and Chen, Tong},
  booktitle={Proceedings of the 33rd ACM International Conference on Multimedia},
  pages={13972--13978},
  year={2025}
}

@inproceedings{lian2025ov,
  title={OV-MER: Towards Open-Vocabulary Multimodal Emotion Recognition},
  author={Lian, Zheng and Sun, Haiyang and Sun, Licai and Chen, Haoyu and Chen, Lan and Gu, Hao and Wen, Zhuofan and Chen, Shun and Siyuan, Zhang and Yao, Hailiang and others},
  booktitle={International Conference on Machine Learning},
  pages={37015--37050},
  year={2025},
  organization={PMLR}
}

@article{li2025multimodal,
  title={Multimodal PEAR chain-of-thought reasoning for multimodal sentiment analysis},
  author={Li, Yan and Lan, Xiangyuan and Chen, Haifeng and Lu, Ke and Jiang, Dongmei},
  journal={ACM Transactions on Multimedia Computing, Communications and Applications},
  volume={20},
  number={9},
  pages={1--23},
  year={2025},
  publisher={ACM New York, NY}
}

@inproceedings{seoh2025emogist,
  title={EmoGist: Efficient In-Context Learning for Visual Emotion Understanding},
  author={Seoh, Ronald and Goldwasser, Dan},
  booktitle={Findings of the Association for Computational Linguistics: EMNLP 2025},
  pages={2171--2182},
  year={2025}
}

@inproceedings{fang2025catch,
  title={Catch your emotion: Sharpening emotion perception in multimodal large language models},
  author={Fang, Yiyang and Liang, Jian and Huang, Wenke and Li, He and Su, Kehua and Ye, Mang},
  booktitle={Forty-second International Conference on Machine Learning},
  year={2025}
}

@inproceedings{yang2025mermaid,
  title={MERMAID: Multi-perspective Self-reflective Agents with Generative Augmentation for Emotion Recognition},
  author={Yang, Zhongyu and Song, Junhao and Song, Siyang and Pang, Wei and Yuan, Yingfang},
  booktitle={Proceedings of the 2025 Conference on Empirical Methods in Natural Language Processing},
  pages={24650--24666},
  year={2025}
}

@inproceedings{tang2025reason,
  title={Reason-before-retrieve: One-stage reflective chain-of-thoughts for training-free zero-shot composed image retrieval},
  author={Tang, Yuanmin and Zhang, Jue and Qin, Xiaoting and Yu, Jing and Gou, Gaopeng and Xiong, Gang and Lin, Qingwei and Rajmohan, Saravan and Zhang, Dongmei and Wu, Qi},
  booktitle={Proceedings of the Computer Vision and Pattern Recognition Conference},
  pages={14400--14410},
  year={2025}
}

@inproceedings{wangvideorft,
  title={VideoRFT: Incentivizing Video Reasoning Capability in MLLMs via Reinforced Fine-Tuning},
  author={Wang, Qi and Yu, Yanrui and Yuan, Ye and Mao, Rui and Zhou, Tianfei},
  booktitle={The Thirty-ninth Annual Conference on Neural Information Processing Systems}
}

@inproceedings{wang2025embracing,
  title={Embracing Collaboration Over Competition: Condensing Multiple Prompts for Visual In-Context Learning},
  author={Wang, Jinpeng and Luo, Tianci and Zha, Yaohua and Feng, Yan and Luo, Ruisheng and Chen, Bin and Dai, Tao and Chen, Long and Wang, Yaowei and Xia, Shu-Tao},
  booktitle={Proceedings of the Computer Vision and Pattern Recognition Conference},
  pages={25156--25165},
  year={2025}
}

@inproceedings{fei2024video,
  title={Video-of-Thought: Step-by-Step Video Reasoning from Perception to Cognition},
  author={Fei, Hao and Wu, Shengqiong and Ji, Wei and Zhang, Hanwang and Zhang, Meishan and Lee, Mong-Li and Hsu, Wynne},
  booktitle={International Conference on Machine Learning},
  pages={13109--13125},
  year={2024},
  organization={PMLR}
}

@inproceedings{ye2024mplug,
  title={mplug-owl2: Revolutionizing multi-modal large language model with modality collaboration},
  author={Ye, Qinghao and Xu, Haiyang and Ye, Jiabo and Yan, Ming and Hu, Anwen and Liu, Haowei and Qian, Qi and Zhang, Ji and Huang, Fei},
  booktitle={Proceedings of the ieee/cvf conference on computer vision and pattern recognition},
  pages={13040--13051},
  year={2024}
}

@inproceedings{mitra2024compositional,
  title={Compositional chain-of-thought prompting for large multimodal models},
  author={Mitra, Chancharik and Huang, Brandon and Darrell, Trevor and Herzig, Roei},
  booktitle={Proceedings of the IEEE/CVF Conference on Computer Vision and Pattern Recognition},
  pages={14420--14431},
  year={2024}
}

@inproceedings{fang2025emoe,
  title={Emoe: Modality-specific enhanced dynamic emotion experts},
  author={Fang, Yiyang and Huang, Wenke and Wan, Guancheng and Su, Kehua and Ye, Mang},
  booktitle={Proceedings of the Computer Vision and Pattern Recognition Conference},
  pages={14314--14324},
  year={2025}
}

@inproceedings{yang2025mse,
  title={Mse-adapter: A lightweight plugin endowing llms with the capability to perform multimodal sentiment analysis and emotion recognition},
  author={Yang, Yang and Dong, Xunde and Qiang, Yupeng},
  booktitle={Proceedings of the AAAI Conference on Artificial Intelligence},
  volume={39},
  number={24},
  pages={25642--25650},
  year={2025}
}

@inproceedings{zhang2024visual,
  title={Visual prompting in LLMs for enhancing emotion recognition},
  author={Zhang, Qixuan and Wang, Zhifeng and Zhang, Dylan and Niu, Wenjia and Caldwell, Sabrina and Gedeon, Tom and Liu, Yang and Qin, Zhenyue},
  booktitle={Proceedings of the 2024 Conference on Empirical Methods in Natural Language Processing},
  pages={4484--4499},
  year={2024}
}

@inproceedings{panda2018contemplating,
  title={Contemplating visual emotions: Understanding and overcoming dataset bias},
  author={Panda, Rameswar and Zhang, Jianming and Li, Haoxiang and Lee, Joon-Young and Lu, Xin and Roy-Chowdhury, Amit K},
  booktitle={Proceedings of the European Conference on Computer Vision (ECCV)},
  pages={579--595},
  year={2018}
}

@inproceedings{yang2023emoset,
  title={Emoset: A large-scale visual emotion dataset with rich attributes},
  author={Yang, Jingyuan and Huang, Qirui and Ding, Tingting and Lischinski, Dani and Cohen-Or, Danny and Huang, Hui},
  booktitle={Proceedings of the IEEE/CVF International Conference on Computer Vision},
  pages={20383--20394},
  year={2023}
}

@inproceedings{peng2015mixed,
  title={A mixed bag of emotions: Model, predict, and transfer emotion distributions},
  author={Peng, Kuan-Chuan and Chen, Tsuhan and Sadovnik, Amir and Gallagher, Andrew C},
  booktitle={Proceedings of the IEEE conference on computer vision and pattern recognition},
  pages={860--868},
  year={2015}
}

@inproceedings{machajdik2010affective,
  title={Affective image classification using features inspired by psychology and art theory},
  author={Machajdik, Jana and Hanbury, Allan},
  booktitle={Proceedings of the 18th ACM international conference on Multimedia},
  pages={83--92},
  year={2010}
}

@article{fang2026emo,
  title={EMO-R3: Reflective Reinforcement Learning for Emotional Reasoning in Multimodal Large Language Models},
  author={Fang, Yiyang and Huang, Wenke and Fu, Pei and Yang, Yihao and Su, Kehua and Luo, Zhenbo and Luan, Jian and Ye, Mang},
  journal={arXiv preprint arXiv:2602.23802},
  year={2026}
}

@inproceedings{weng2023affective,
  title={Affective image filter: Reflecting emotions from text to images},
  author={Weng, Shuchen and Zhang, Peixuan and Chang, Zheng and Wang, Xinlong and Li, Si and Shi, Boxin},
  booktitle={Proceedings of the IEEE/CVF International Conference on Computer Vision},
  pages={10810--10819},
  year={2023}
}

@inproceedings{you2016building,
  title={Building a large scale dataset for image emotion recognition: The fine print and the benchmark},
  author={You, Quanzeng and Luo, Jiebo and Jin, Hailin and Yang, Jianchao},
  booktitle={Proceedings of the AAAI conference on artificial intelligence},
  volume={30},
  number={1},
  year={2016}
}

@article{wah2011caltech,
  title={The caltech-ucsd birds-200-2011 dataset},
  author={Wah, Catherine and Branson, Steve and Welinder, Peter and Perona, Pietro and Belongie, Serge},
  year={2011}
}

@inproceedings{krause20133d,
  title={3d object representations for fine-grained categorization},
  author={Krause, Jonathan and Stark, Michael and Deng, Jia and Fei-Fei, Li},
  booktitle={Proceedings of the IEEE international conference on computer vision workshops},
  pages={554--561},
  year={2013}
}

@article{maji2013fine,
  title={Fine-grained visual classification of aircraft},
  author={Maji, Subhransu and Rahtu, Esa and Kannala, Juho and Blaschko, Matthew and Vedaldi, Andrea},
  journal={arXiv preprint arXiv:1306.5151},
  year={2013}
}

@inproceedings{dataset2011novel,
  title={Novel datasets for fine-grained image categorization},
  author={Dataset, E},
  booktitle={First workshop on fine grained visual categorization, CVPR. Citeseer. Citeseer. Citeseer},
  volume={5},
  number={1},
  pages={2},
  year={2011},
  organization={Citeseer}
}

@inproceedings{atabuzzaman2025zero,
  title={Zero-Shot Fine-Grained Image Classification Using Large Vision-Language Models},
  author={Atabuzzaman, Md and Zhang, Andrew and Thomas, Chris},
  booktitle={The 2025 Conference on Empirical Methods in Natural Language Processing},
  year={2025},
}

@inproceedings{kim2024finer,
  title={Finer: Investigating and enhancing fine-grained visual concept recognition in large vision language models},
  author={Kim, Jeonghwan and Ji, Heng},
  booktitle={Proceedings of the 2024 Conference on Empirical Methods in Natural Language Processing},
  pages={6187--6207},
  year={2024}
}

@article{qwen2025qwen25technicalreport,
  title={Qwen2. 5-VL Technical Report},
  author={Bai, Shuai and Chen, Keqin and Liu, Xuejing and Wang, Jialin and Ge, Wenbin and Song, Sibo and Dang, Kai and Wang, Peng and Wang, Shijie and Tang, Jun and others},
  journal={arXiv preprint arXiv:2502.13923},
  year={2025}
}

@book{parrott2001emotions,
  title={Emotions in social psychology: Essential readings},
  author={Parrott, W Gerrod},
  year={2001},
  publisher={psychology press}
}

@article{bai2025qwen3,
  title={Qwen3-vl technical report},
  author={Bai, Shuai and Cai, Yuxuan and Chen, Ruizhe and Chen, Keqin and Chen, Xionghui and Cheng, Zesen and Deng, Lianghao and Ding, Wei and Gao, Chang and Ge, Chunjiang and others},
  journal={arXiv preprint arXiv:2511.21631},
  year={2025}
}

@article{wang2025internvl3,
  title={Internvl3. 5: Advancing open-source multimodal models in versatility, reasoning, and efficiency},
  author={Wang, Weiyun and Gao, Zhangwei and Gu, Lixin and Pu, Hengjun and Cui, Long and Wei, Xingguang and Liu, Zhaoyang and Jing, Linglin and Ye, Shenglong and Shao, Jie and others},
  journal={arXiv preprint arXiv:2508.18265},
  year={2025}
}

@inproceedings{ye2024dual,
  title={Dual-path collaborative generation network for emotional video captioning},
  author={Ye, Cheng and Chen, Weidong and Li, Jingyu and Zhang, Lei and Mao, Zhendong},
  booktitle={Proceedings of the 32nd ACM International Conference on Multimedia},
  pages={496--505},
  year={2024}
}

@article{ye2025improving,
  title={Improving video summarization by exploring the coherence between corresponding captions},
  author={Ye, Cheng and Chen, Weidong and Hu, Bo and Zhang, Lei and Zhang, Yongdong and Mao, Zhendong},
  journal={IEEE Transactions on Image Processing},
  year={2025},
  publisher={IEEE}
}

@inproceedings{ye2025multi,
  title={Multi-round mutual emotion-cause pair extraction for emotion-attributed video captioning},
  author={Ye, Cheng and Chen, Weidong and Song, Peipei and Liu, Xinyan and Zhang, Lei and Mao, Zhendong},
  booktitle={Proceedings of the 33rd ACM International Conference on Multimedia},
  pages={3320--3329},
  year={2025}
}

@article{chen2026subjective,
  title={Subjective-objective emotion correlated generation network for subjective video captioning},
  author={Chen, Weidong and Ye, Cheng and Song, Peipei and Zhang, Lei and Zhang, Yongdong and Mao, Zhendong},
  journal={IEEE Transactions on Image Processing},
  year={2026},
  publisher={IEEE}
}

@article{hong2026emostyle,
  title={EmoStyle: Affective Conditioning of Style-Specialist Experts for Emotional Image Generation},
  author={Hong, Dexiang and Guo, Yijie and Chen, Weidong and Liu, Xinyan and Zou, Zixuan and Mao, Zhendong and Zhang, Yongdong},
  journal={arXiv preprint arXiv:2607.10165},
  year={2026}
}

@article{chen2026creatiparser,
  title={Creatiparser: Generative image parsing of raster graphic designs into editable layers},
  author={Chen, Weidong and Hong, Dexiang and Mao, Zhendong and Cheng, Yutao and Liu, Xinyan and Zhang, Lei and Zhang, Yongdong},
  journal={arXiv preprint arXiv:2604.19632},
  year={2026}
}

@inproceedings{huang2025graph,
  title={Graph mixture of experts and memory-augmented routers for multivariate time series anomaly detection},
  author={Huang, Xiaoyu and Chen, Weidong and Hu, Bo and Mao, Zhendong},
  booktitle={Proceedings of the AAAI conference on artificial intelligence},
  volume={39},
  number={16},
  pages={17476--17484},
  year={2025}
}

@article{song2025towards,
  title={Towards efficient partially relevant video retrieval with active moment discovering},
  author={Song, Peipei and Zhang, Long and Lan, Long and Chen, Weidong and Guo, Dan and Yang, Xun and Wang, Meng},
  journal={IEEE Transactions on Multimedia},
  year={2025},
  publisher={IEEE}
}

@inproceedings{chen2022multi,
  title={Multi-attention network for compressed video referring object segmentation},
  author={Chen, Weidong and Hong, Dexiang and Qi, Yuankai and Han, Zhenjun and Wang, Shuhui and Qing, Laiyun and Huang, Qingming and Li, Guorong},
  booktitle={Proceedings of the 30th ACM international conference on multimedia},
  pages={4416--4425},
  year={2022}
}

@inproceedings{wang2023improving,
  title={Improving image captioning via predicting structured concepts},
  author={Wang, Ting and Chen, Weidong and Tian, Yuanhe and Song, Yan and Mao, Zhendong},
  booktitle={Proceedings of the 2023 conference on empirical methods in natural language processing},
  pages={360--370},
  year={2023}
}

@inproceedings{chen2021cascade,
  title={Cascade cross-modal attention network for video actor and action segmentation from a sentence},
  author={Chen, Weidong and Li, Guorong and Zhang, Xinfeng and Yu, Hongyang and Wang, Shuhui and Huang, Qingming},
  booktitle={Proceedings of the 29th ACM International Conference on Multimedia},
  pages={4053--4062},
  year={2021}
}

@article{chen2023weakly,
  title={Weakly supervised text-based actor-action video segmentation by clip-level multi-instance learning},
  author={Chen, Weidong and Li, Guorong and Zhang, Xinfeng and Wang, Shuhui and Li, Liang and Huang, Qingming},
  journal={ACM Transactions on Multimedia Computing, Communications and Applications},
  volume={19},
  number={1},
  pages={1--22},
  year={2023},
  publisher={ACM New York, NY}
}
